\documentclass[10pt,twocolumn,letterpaper]{article}

\usepackage[pagenumbers]{iccv} % To force page numbers, e.g. for an arXiv version
\makeatletter
\@namedef{ver@everyshi.sty}{}
\makeatother
\usepackage{graphicx}
\usepackage{booktabs}
\usepackage{tikz}
\usepackage{siunitx}
\usepackage{verbatim}
\usepackage{color}
\usepackage{colortbl}
\usepackage{booktabs}
\usepackage{multirow}
\usepackage{makecell}
\usepackage{pifont} 
\usepackage{soul}
\usepackage{listings}
\usepackage{rotating}
\usepackage{tikz,pgf}
\usepackage{tkz-euclide}
\usepackage{pgfplots}

\newcommand*\rot{\rotatebox{90}}

\definecolor{mygray}{gray}{.92}
\newsavebox\CBox

\newcommand{\M}[1]{\mathbf{#1}}
\definecolor{wincolor}{rgb}{0.95, 0.2, 0.2}

\newcommand{\ScanNet}{\dataset{ScanNet}}
\newcommand{\ETH}{\dataset{ETH3D}}

\newcommand{\Phototourism}{\dataset{Phototourism}}

\DeclareMathSymbol{\shortminus}{\mathbin}{AMSa}{"39}
\newcommand{\dataset}[1]{{\fontfamily{cmtt}\selectfont #1} }

\usepackage[dvipsnames]{xcolor}
\newcommand{\red}[1]{{\color{red}#1}}
\newcommand{\blue}[1]{{\color{blue}#1}}

\pgfplotsset{
compat=1.9,
legend image code/.code={
\draw[mark repeat=4,mark phase=1]
plot coordinates {
(0cm,0.02cm)
(0.1cm,0.02cm)        
(0.2cm,0.02cm)
(0.3cm,0.02cm)       
};%
}
}

\definecolor{Seaborn1}{HTML}{1f77b4}
\definecolor{Seaborn2}{HTML}{ff7f0e}
\definecolor{Seaborn3}{HTML}{2ca02c}
\definecolor{Seaborn4}{HTML}{d62728}
\definecolor{Seaborn5}{HTML}{9467bd}
\definecolor{Seaborn6}{HTML}{8c564b}
\definecolor{Seaborn7}{HTML}{e377c2}
\definecolor{Seaborn8}{HTML}{7f7f7f}
\definecolor{Seaborn9}{HTML}{bcbd22}
\definecolor{Seaborn10}{HTML}{17becf}

\definecolor{iccvblue}{rgb}{0.21,0.49,0.74}
\usepackage[pagebackref,breaklinks,colorlinks,citecolor=iccvblue]{hyperref}

\def\paperID{6260} % *** Enter the Paper ID here
\def\confName{ICCV}
\def\confYear{2025}

\title{RePoseD: Efficient Relative Pose Estimation With Known Depth Information}

\author{Yaqing Ding$^{1}$, Viktor Kocur$^2$, Václav Vávra$^1$, Zuzana Berger Haladová$^2$, Jian Yang$^3$,\\ Torsten Sattler$^4$ and Zuzana Kukelova$^1$\\
	$^1$ Visual Recognition Group, Faculty of Electrical Engineering, 
	Czech Technical University in Prague \\
    $^2$ Faculty of Mathematics, Physics and Informatics, Comenius University in Bratislava\\
    $^3$ VCIP, CS, Nankai University, Tianjin, China\\
    $^4$ Czech Institute of Informatics, Robotics and Cybernetics, Czech Technical University in Prague
	\\
}

\begin{document}
\maketitle
\begin{abstract}
% Recent advances in depth prediction have led to significantly improved depth prediction accuracy. In turn, this enables various applications to use such depth predictions. In this paper, we propose a novel framework for estimating the relative pose between two cameras from point correspondences with associated monocular depths. 
% Since depth predictions are typically defined up to an unknown scale or even both scale and shift parameters, our solvers jointly estimate the scale or both the scale and shift parameters along with the relative pose. We derive efficient solvers considering different types of depths for three camera configurations: (1) two calibrated cameras, (2) two uncalibrated cameras with an unknown but shared focal length, and (3) two uncalibrated cameras with unknown and different focal lengths. Experiments on three large-scale real-world datasets with depth maps from five different depth methods show the practical viability of our solvers. Compared to previous work, we propose more efficient solvers that achieve state-of-the-art results. The code will be made publicly available. 

Recent advances in monocular depth estimation methods (MDE) and their improved accuracy open new possibilities for their applications. 
In this paper, we investigate how monocular depth estimates can be used for relative pose estimation. 
In particular, we are interested in answering the question whether using MDEs improves  results over traditional point-based methods. 
We propose a novel framework for estimating the relative pose of two cameras from point correspondences with associated monocular depths. 
Since depth predictions are typically defined up to an unknown scale or even both unknown scale and shift parameters, our solvers jointly estimate the scale or both the scale and shift parameters along with the relative pose. We derive efficient solvers considering different types of depths for three camera configurations: (1) two calibrated cameras, (2) two cameras with an unknown shared focal length, and (3) two cameras with unknown different focal lengths. 
Our new solvers outperform state-of-the-art depth-aware solvers in terms of speed and accuracy. In extensive real experiments on multiple datasets and with various MDEs, we discuss which depth-aware solvers are preferable in which situation. 
The code will be made publicly available. 
%Experiments on three large-scale real-world datasets with depth maps from five different depth methods show the practical viability of our solvers. Compared to previous work, we propose more efficient solvers that achieve state-of-the-art results. The code will be made publicly available. 
\end{abstract}  
\section{Introduction}
Relative pose estimation is a core problem in a wide range of computer vision applications, including Structure-from-Motion (SfM)~\cite{schonberger2016structure}, visual localization~\cite{zeisl2015camera,svarm2016city,sarlin2019coarse}, and autonomous robots~\cite{Lim2015IJRR}. 
Typically, the relative pose of cameras is estimated from 2D-2D image point correspondences that are established using feature matching. Given 2D-2D matches, the relative pose~\cite{nister2004efficient,stewenius2005minimal,hartley2012efficient,kukelova2012polynomial,kukelova2017clever,stewenius2006recent} of two cameras can be estimated using the epipolar geometry constraints~\cite{hartley2003multiple}. For two calibrated cameras, 5 %image 
point correspondences are needed to estimate the 5 degrees of freedom of the relative pose (3 for rotation and two for the translation up to an unknown scale) 
%using the well-known 5-point algorithm
~\cite{nister2004efficient,hartley2012efficient,kukelova2012polynomial,stewenius2006recent}. 
%(or using 4 point correspondences to estimate a homography~\cite{hartley2003multiple} in the case of planar scenes). 
%If the focal length is unknown, but both cameras share the same focal length, the relative pose can be estimated together with the focal length from 6 point correspondences~\cite{stewenius2005minimal,hartley2012efficient,kukelova2012polynomial,kukelova2017clever}. 
The relative pose of two cameras with unknown shared focal length can be estimated from 6 %point 
correspondences~\cite{stewenius2005minimal,hartley2012efficient,kukelova2012polynomial,kukelova2017clever}. 
If the full intrinsic calibration is unknown and/or both cameras have different calibrations, the relative pose %, together with the calibration matrices for the camers, 
can be estimated from 7 or 8  correspondences~\cite{hartley2003multiple}.  %7- or 8-point correspondences~\cite{hartley2003multiple}. 

Due to noise in the measurements and the presence of outlier correspondences, algorithms for relative pose estimation are usually employed in RANSAC-style hypothesize-and-verify frameworks~\cite{fischler1981random}.
%in order to handle outlier correspondences. 
The number of RANSAC iterations grows exponentially with 
%the outlier ratio, \ie, the ratio of incorrect correspondences among all correspondences, and 
the number of correspondences needed to estimate the pose. 
As such, a considerable amount of work focuses on reducing the number of correspondences necessary for pose estimation. This is typically done by using additional information, such as the gravity direction from an Inertial Measurement Unit (IMU)~\cite{fraundorfer2010minimal,naroditsky2012two,kukelova2010closed,sweeney2014solving}, or information about the local feature geometry such as the scale and rotation of features~\cite{barath2022relative,liwicki2017scale,guan2022relative} or local affine frames~\cite{Bentolila2014,barath2017minimal,eichhardt2020relative}.

\begin{figure}[t]
    \centering
    \includegraphics[width=0.99\linewidth]{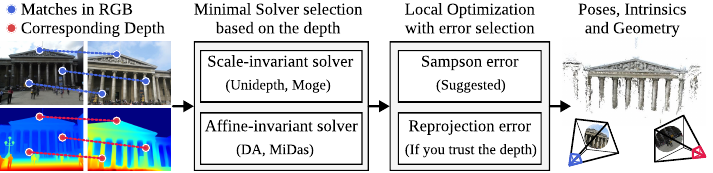}
    \caption{Given point matches and their corresponding depths from a pair of images, we propose the use of different minimal solvers for relative pose estimation, according to the properties of the depth data. For local optimization, the Sampson error is generally more robust across varying depth conditions, while the reprojection error may yield better results when the depth measurements are highly reliable.}
    \label{fig:teaser}
    \vspace{-5mm}
\end{figure}

Recently, learning-based methods~\cite{godard2019digging,yang2024depth,wang2024moge,hu2024metric3d,bochkovskii2024depth,piccinelli2024unidepth,martingarcia2024diffusione2eft,yang2024depthv2} have achieved significant success in predicting depth from a single image. The increasing precision of Monocular Depth Estimation methods (MDE) opens new possibilities of using depth information in downstream tasks.
Recent studies have shown a growing focus on the use of dense depth estimates to improve the performance of various geometry-based applications, including structure-from-motion/SLAM~\cite{merrill2023fast,zhou2022learned,liu2022depth}, dense reconstruction~\cite{yu2022monosdf,yang2024gaussianobject}, and novel view synthesis~\cite{jiang2024construct,li2024dngaussian}. 

%Monocular 
Depth naturally provides additional information for camera pose estimation.  Thus, several recent works~\cite{Astermark2024,Barathicpr} attempted to use monocular depths from MDE networks or networks to estimate  the relative depths of two features~\cite{dingfundamental} to improve the relative pose estimation of cameras. The depth information provides additional constraints on the geometry, and thus reduces the number of correspondences necessary for the estimation.
However, it also introduces new challenges: 
(1) Learning-based depth estimates are usually much more noisy and erroneous than image measurements. (2) 
%and in this way improve the performance of relative pose solvers.
%However, very few studies have attempted to improve the relative pose estimation using monocular depth. 
%Learning-based depth estimation
MDE methods typically produce scale/affine-invariant depth~\cite{wang2024moge,martingarcia2024diffusione2eft,ke2023repurposing,bochkovskii2024depth}, affine-invariant inverse depth~\cite{yang2024depth,yang2024depthv2,birkl2023midas,ranftl2021vision} or metric depth~\cite{hu2024metric3d,bhat2023zoedepth,piccinelli2024unidepth}, meaning that the estimated depth or inverse depth is %can be defined 
only defined up to an unknown scale factor or both unknown scale and shift parameters. (3) The unknown shift and scale can be different for different images of the same scene. 
The scale may not even be consistent across an image, \ie, some parts of the image may be scaled by a different scale than others.
Due to these challenges and the fact that the solvers proposed in~\cite{Astermark2024,dingfundamental} use only relative depths, %they 
do not model unknown shifts, and have been tested %the authors test their methods 
only with depths from one network (Depth Anything~\cite{yang2024depthv2} in~\cite{dingfundamental}, and a network for estimating the relative depth proposed directly in~\cite{Astermark2024}), 
%the papers~
\cite{Astermark2024,dingfundamental} do not convincingly %were unable to convincingly 
show the improvement of their 
%new
depth-aware solvers over traditional 2D-2D point-based relative pose solvers~\cite{nister2004efficient,hartley2003multiple,stewenius2006recent}.
Thus, prior work does not answer the question of whether recent MDE networks provide useful information for the relative pose estimation task.

This paper attempts to provide an answer to this question by presenting a comprehensive comparison of different depth-aware solvers for calibrated, shared focal length and different focal lengths cases, under different conditions, \ie, different sources of depth information, different feature detectors and matchers, different types of datasets (interior/exterior) and different optimization schemes inside RANSAC. 
%In this paper, inspired by advancements in monocular depth estimation, this paper presents a comprehensive discussion on relative pose estimation using monocular depth in two cameras.
To this end, we provide novel solvers 
%Specifically, we consider three categories of 
for three different types of monocular depths: metric depth, scale-invariant depth, and affine-invariant depth. We introduce novel formulations that jointly estimate the relative pose of two cameras along with the scale or affine parameters of the monocular depths. The proposed solvers use the full information from the depth estimates, resulting in fewer correspondences necessary for the relative pose estimation compared to %the 
solvers based on relative depths~\cite{dingfundamental}. 
%In extensive experiments 
%We demonstrate that integrating depth information with relative pose estimation not only enhances efficiency but also improves accuracy. 

The main contributions of the paper are:
\begin{itemize}
  \item[$\bullet$] We integrate scale and affine parameters of monocular depth into solvers for relative pose estimation. These parameters are estimated together with the relative pose. %, thus formulating a problem of their joint estimation.
  \item[$\bullet$] We discuss all possible combinations of known and unknown shift and scale parameters. We propose several practical and efficient minimal solvers for calibrated cameras, two cameras with shared unknown focal length, and two cameras with different unkown focal lengths. 
  Our %The proposed 
  solvers outperform state-of-the-art depth-aware solvers in terms of efficiency and accuracy.
  \item[$\bullet$] We evaluate different RANSAC schemes to account for potentially erroneous and noisy depth estimates.
%  \item[$\bullet$] In extensive real and synthetic experiments, we show an improvement in terms of accuracy and speed over the existing solvers. 
 \item[$\bullet$] In extensive real experiments, we answer the questions which %popular 
 state-of-the-art MDE networks are suitable for relative pose estimation, which %of them 
 require modeling unknown scale or affine parameters, and which provide the best improvement in terms of efficiency and accuracy over %standard %2d-2d 
 traditional, purely point correspondence-based solvers. % based purely on point correspondences. %based solvers when used for the relative pose estimation.
 %inside the new proposed solvers.
\end{itemize}

\section{Related Work}

% Several previous studies have explored the use of depth priors for relative pose estimation. In~\cite{liwicki2017scale}, it has been shown that the relative scales of the SIFT~\cite{lowe2004distinctive} features can be approximated as the inverse relative depths of the features. 
\cite{liwicki2017scale} shown that the relative scales of  SIFT~\cite{lowe2004distinctive} features can %be 
approximate %as 
the inverse relative depths of the features. 
In this case, one point correspondence provides two constraints for relative pose estimation, and the 5-DOF problem can be solved using only three point correspondences with two relative depths. 
% With this assumption, 
In the same setting, Guan~\etal~\cite{guan2022relative} solve the generalized relative pose problem using three SIFT correspondences. Astermark~\etal~\cite{Astermark2024} propose a closed-form minimal solution to the relative pose of two calibrated pinhole cameras with known relative depth. In addition, they propose a neural network to estimate the relative depth. 
%In~\cite{ding2024noisy}, the relative depth prior is used for homography estimation. %They are mainly focus on the calibrated case using relative depths approximated from SIFT scales. 
In~\cite{ding2024noisy}, relative depths approximated from SIFT scales are used to estimate the calibrated homography.
Barath~\etal~\cite{Barathicpr} propose a 2-point solver with depth information for relative pose estimation. However, as pointed out in~\cite{yu2025relative}, this 2-point solver is degenerate for rigid alignments due to rank deficiency. 
%In~\cite{dingfundamental}, the authors assume that the shift in monocular depth can be ignored and propose new minimal solvers that use relative depth for the fundamental matrix estimation with different variants of known and unknown focal lengths. Their solvers are unaffected by the scale of the monocular depth. 
In~\cite{dingfundamental}, the authors propose new minimal solvers that use relative depth for fundamental matrix estimation. They discuss different variants of solvers with known and unknown focal lengths. 
Their %proposed 
solvers work for monocular depths with unknown different scales. However, the solvers do not model unknown shifts in the depths, assuming that the shifts are zero in both images and thus can be ignored.
All of the methods and minimal solvers discussed above use only the information from relative depths. Thus, they do not utilize the full information that two monocular depth maps can provide.

% Recently, 
Dust3r~\cite{wang2024dust3r} is a recent %proposed a 
3D foundation model for %designed for 
two-view geometry estimation tasks, including depth prediction and relative pose estimation. 
% Leveraging large-scale training to achieve significant advances. 
The relative pose is estimated by aligning two point maps with a similarity transformation. 
Trained on large-scale datasets, Dust3r is able to handle scenarios where there is little overlap between the two images. 
% Furthermore, 
Mast3r~\cite{leroy2024grounding} further improves performance through %enhanced this approach by utilizing 
feature matching. 
Mast3r %and using PyTorch to 
minimizes both 3D-3D matching errors and 3D-2D reprojection errors to refine %obtain the final 
depths, relative poses, and intrinsic parameters.

In this paper, in contrast to the minimal solvers mentioned above that use only relative depths, we solve the problem using the full information from two depth maps. This allows us to formulate the relative pose problem using different combinations of 3D-3D/3D-2D/2D-2D correspondences.
%\footnote{The 2D image points that are equipped with depths can be lifted to 3D and considered as 3D points.}. 
One 3D-3D correspondence provides three constraints and allows us to solve the relative pose problem from fewer correspondences than~\cite{dingfundamental}. Moreover, 
%in contrast to previous work, 
we propose new minimal solvers that model not only unknown scales, but also unknown shifts in depth maps. 
\vspace{-3ex}
\paragraph{Concurrently to our work,} %Madpose: 
Yu \etal~\cite{yu2025relative} %concurrently 
developed three novel minimal relative pose solvers for affine-invariant depths. 
Their % proposed 
calibrated, shared-focal, and two-focal length solvers solve the same formulation as our solvers with unknown shift and scales. In order to deal with potentially noisy and erroneous depth information, Yu \etal %the authors of~\cite{yu2025relative} 
propose to %jointly 
use their depth-aware solvers
jointly with %the 
classical 2D-2D point-based solvers in a hybrid LO-MSAC~\cite{torr2000mlesac,Camposeco2018CPVR} framework.
In their Madpose framework, they combine the symmetric depth-induced reprojection error and the Sampson error for scoring pose hypotheses and for local optimization.
The main differences between our work and~\cite{yu2025relative} are the following:
(1) Our solutions to configurations with unknown shifts and scales, \ie, affine-invariant depths, result in smaller and thus faster solvers. 
%For example, for the calibrated case, our solution directly leads to finding the roots of a quartic equation, while the solver proposed in~\cite{yu2025relative} performs the Gauss-Jordan elimination of a $12\times 16$ matrix and the eigenvalue decomposition of a $4\times 4$ matrix.
For a detailed comparison of sizes and run-times of the proposed and the Madpose solvers~\cite{yu2025relative}, see Table~\ref{table:eff}. (2) In addition to affine-invariant solvers, we also derive novel minimal solvers for cases with known/zero shifts and either unknown or known/same scales. 
%and known shifts and known/same scales.
Moreover, we derive a calibrated solver for inverse depth and discuss the number of solutions for all other variants of known/unknown/same/different shifts and scales (see Table~\ref{tab:solvers}). (3) By testing more variants of solvers, including novel solvers with zero shifts, 
%and with a wider range of MDE networks, 
we show that in some scenarios it is not necessary to model unknown shifts. This is in contrast to the observation made by~\cite{yu2025relative}, who, by missing some variants of the solvers in their experiments, concluded that modeling the unknown shift is always beneficial %brings the benefit 
(even for ``metric" depths). (4) In contrast to using a complex hybrid RANSAC scheme and a combination of reprojection and Sampson errors that results in slower running times, we look at a simple and more efficient local optimization scheme based on using a single solver with either reprojection or Sampson error. Even for 3D points, we propose to compose the essential/fundamental matrix and measure the Sampson error, as this % that 
usually leads to better results.

\section{Problem Statement}

Assume that a set of 3D points $\Omega$ is observed by two cameras with projection matrices $\M K_1[\M I\ |\ \M 0]$ and $\M K_2[\M R\ |\ \M T]$. Let $\{\M p_{i},\M q_{i}\}  ,i=1,\dots,n$ be a set of $n$ 2D point correspondences, \ie, the projections of 3D points $\Omega$ in the first and the second camera, respectively. Then we have
\begin{equation}
 \lambda_{i}\M K_2^{-1} \M q_{i} =\eta_{i} \M R\M K_1^{-1} \M p_{i}  +\M T, \label{eq:01}
\end{equation}
where $\lambda_{i}$ and $\eta_{i}$ are the depths of the points $\M q_{i}$ and $\M p_{i}$.
%, respectively. 

%Recently, several learning-based methods~\cite{godard2019digging,yang2024depth} that provide non-metric monocular depth estimates, \ie, the depth estimates with unknown scale factor and shift, have been proposed. Using these methods, we can obtain a depth image corresponding to the RGB image. 

%A depth image can provide additional information for relative pose estimation. In this paper, we study, how such monocular depth known up to scale and shift can be used in pose estimation of two cameras. Our goal is to estimate the scale, with or without the shift, along with the motion parameters, simultaneously.

%Modern learning based monocular depth estimation can provide affine invariant depth~\cite{wang2024moge,martingarcia2024diffusione2eft,ke2023repurposing} or affine invariant inverse depth~\cite{yang2024depth,yang2024depthv2}. In the main paper, we only discuss the case where the depth up to scale and shift, and in the supplementary material (SM) we show the case where the inverse depth up to scale and shift. 

MDE networks provide either scale/affine-invariant depth~\cite{wang2024moge,martingarcia2024diffusione2eft,ke2023repurposing,bochkovskii2024depth}, affine-invariant inverse depth~\cite{yang2024depth,yang2024depthv2,birkl2023midas,ranftl2021vision}, or metric depth~\cite{hu2024metric3d,bhat2023zoedepth,piccinelli2024unidepth}. In most general case of affine-invariant depth the true depths $\lambda_i$ and $\eta_i$ in~\eqref{eq:01} can be expressed as
\begin{equation}
% \begin{split}
\eta_i  = s_1(\alpha_{i} + u),\ \lambda_i  = s_2(\beta_{i} + v) , \label{eq:02}
% \end{split}
\end{equation}
where $\alpha_{i}, \beta_{i}$ are estimated affine-invariant depths,
%up to scale and shift, 
and $\{s_1,s_2\},\{u,v\}$ are the unknown scales and shifts
of the depths. Substituting~\eqref{eq:02} into~\eqref{eq:01}, we have
% \begin{equation}
%  s_2(\beta_{i} + v) \tilde{\M q}_i =s_1(\alpha_{i} + u) \M R \tilde{\M p}_i  +\M T, \label{eq:03}
% \end{equation}
\begin{equation}
 s_2(\beta_{i} + v) \M K_2^{-1} \M q_{i} =s_1(\alpha_{i} + u) \M R \M K_1^{-1} \M p_{i}  +\M T, \label{eq:03}
\end{equation}
Dividing~\eqref{eq:03} by $s_1$, results in
\begin{equation}
 s(\beta_{i} + v)\M K_2^{-1} \M q_{i}  =(\alpha_{i} +u) \M R \M K_1^{-1} \M p_{i}  +{\M t}, \label{eq:04}
\end{equation}
where $s = s_2/s_1$ and ${\M t} = {\M T}/s_1$, \ie two different scales are replaced by the relative scale $s$. 
%Note that the scale of the translation has been fixed in this case. Although it is defined up to a scale factor, it retains 3 degrees of freedom (3-DOF). 
Note that in this case, the scale of the translation $\M t$ is defined \wrt $s_1$, thus we can not further fix the scale of $\M t$, i.e the translation vector is not estimated up to scale like in the standard relative pose solvers~\cite{nister2004efficient}, but needs to be estimated as a 3 degrees of freedom (3-DOF) vector.
This means that for calibrated cameras, \ie, for known calibration matrices $\M K_1$ and $\M K_2$, we have a 9-DOF problem w.r.t. $\{s,u,v,\M R, {\M t}\}$.
Since one 3D-3D point correspondence~\footnote{Note that in this case the 2D points $\{\M p_{i},\M q_{i}\}$ are equiped with depths and thus lifted to 3D points.} provides three constraints of the form~\eqref{eq:04}, we need at least 3 point correspondences with their monocular depths to solve this problem for the calibrated case. For the case of unknown focal lengths $f_1$ and $f_2$ in calibration matrices $\M K_1$ and $\M K_2$, we have 10 and 11 DoF, for cases $f_1=f_2$ and  $f_1\neq f_2$ respectively.

In addition to affine-invariant depths, some MDE networks provide scale-invariant depths or even metric depths; moreover, in some applications the depth can be estimated directly using depth sensors. Thus in some scenarios, we can assume known/zero shift and/or equal scales and shifts of two depths.
%In some scenarios, we may have some additional
%In addition, considering some 
%priors on 
%information about
%the scale and shift $\{s,u,v\}$, \eg, assuming zero shift or equal scale.
There are 6 possible combinations of known and unknown shifts and scales that result in different DOF problems. Table~\ref{tab:solvers} lists all these combinations together with the number of solutions for different problem settings, with novel solvers to interesting practical setups that are tested in the main paper and in the Supplemental Material (SM) highlighted in red and blue, respectively.
Note that for calibrated case the setups with known shifts and scales ($\{s,u,v\} = \{1,0,0\}$), and with known shifts and unknown scale $\{s,u,v\} = \{s,0,0\}$ can be solved using the well-known P3P solver~\cite{gao2003complete,persson2018lambda,ding2023revisiting}. The P3P solver can handle scale invariant depth, since it only uses single side depth.~\footnote{Although one 3D-3D point correspondence provides three constraints, we can not use only two 3D-3D point correspondences to solve 6-DOF pose problem with known $\{s,u,v\}=\{1,0,0\}$. The reason is that the rotation cannot be determined with only two points. There still remains a 1-DOF rotation around the line passing through the two points.}
Due to space limits, we only discuss two practical scenarios in the main paper: the two depths up to unknown scale with two different shifts, and the two depth maps up to unknown scale. Other cases are discussed in the SM.

\begin{table}[!t]
\begin{center}
\resizebox{0.99\linewidth}{!}
        { 
  \begin{tabular}{ccccccccc}
    \toprule
     &  \multirow{2.5}{*}{Solver} &\multicolumn{3}{c}{Point Correspondence}  & \multirow{2.5}{*}{\makecell{ Scale \\ Ratio}} & \multirow{2.5}{*}{Shift} & \multirow{2.5}{*}{DOF} & \multirow{2.5}{*}{\makecell{ No. of \\ Solutions}}  \\ \cmidrule(rl){3-5}    
   &  & 3D-3D & 3D-2D & 2D-2D &  &  &    & \\
      \midrule
 \multirow{8}{*}{\rot{Calibrated}} &  5-point~\cite{nister2004efficient} & - & - & 5  &   - & - & 5&  10  \\
 &  Rel3PT$^*$~\cite{Astermark2024} & 2 & - & 1 & 1 & (0, 0) & 5&  4  \\
 &  P3P~\cite{ding2023revisiting} & - & 3 & -  &   s & (0, -) & 6&  4  \\
 &  3PT$_{1uu}$ & 1 & 2 & 0  &   1 & (u, u) & 7&  8  \\
 &  3PT$_{1uv}$ & 2 & 1 & 0  &   1 & (u, v) & 8&  8  \\
 &  3PT$_{suu}$ & 2 & 1 & 0  &   s & (u, u) & 8& 8  \\
  & \red{3PT$_{suv}$} & 3 & 0 & 0  &   s & (u, v) & 9&  4  \\
  & 3PT$_{suv}$ (inverse) & 3 & 0 & 0  &   s & (u, v) & 9&  10  \\
    \bottomrule
  % & \multirow{2.5}{*}{\makecell{ $\M F$\\ Solver}} &\multicolumn{3}{c}{Point Correspondence}  & \multirow{2.5}{*}{\makecell{ Scale \\ Ratio}} & \multirow{2.5}{*}{Shift} & \multirow{2.5}{*}{DOF} &   \multirow{2.5}{*}{\makecell{ No. of \\ Solutions}}  \\ \cmidrule(rl){3-5}    
  %  &  & 3D-3D & 3D-2D & 2D-2D &  &  &   & \\
  %     \midrule
\multirow{8}{*}{\rot{Equal $f$}} &   6-point~\cite{hartley2012efficient} & - & - & 6 & - & - &  6 & 15  \\
  & 3p3d$^*$~\cite{dingfundamental} & 3 & - & - & 1 &  (0, 0) &  6 & 4  \\
  % & P4Pf~\cite{} & - & 3 & - & 6  &   \textcolor{red}{\ding{55}} & 4 & Closed-form  & 200  \\
  & \blue{3PT$_{100}f$} & 1 & 2 & 0 & 1 &  (0, 0)&  7 & 6   \\
  & 3PT$_{1uu}f$ & 2 & 1 & 0 & 1 &  (u, u)&  8 & 6 \\
  & 3PT$_{1uv}f$ & 3 & 0 & 0 & 1 &  (u, v)&  9 & 4  \\
  & \red{3PT$_{s00}f$} & 2 & 1 & 0 & s &  (0, 0)&  8 & 4  \\
  & 3PT$_{suu}f$ & 3 & 0 & 0 & s &  (u, u)&  9 & 6   \\
  & \red{4PT$_{suv}f$} & 3 & 0 & 1 & s &  (u, v)&  10 & 8   \\
  % & 3PTi & 3 & - & - & 6 &    \textcolor{green}{\ding{51}} & 4 & $4\times 4$  & 106  \\
   % 4PTs && - & 4 & -  && 1 & \textcolor{green}{\ding{51}} && 8 & $8\times 8$  && 106   \\
    \midrule
    \multirow{8}{*}{\rot{Varying $f_1,f_2$}} &   7-point~\cite{hartley2003multiple} & - & - & 7 & - & - & 7 & 3    \\
  & 4p4d$^*$~\cite{dingfundamental} & 4 & - & - & s & (0, 0) & 7 & 1    \\
  & \blue{3PT$_{100}f_{1,2}$} & 2 & 1 & 0 & 1 &  (0, 0) & 8 & 3   \\
  & 3PT$_{1uu}f_{1,2}$ & 3 & 0 & 0 & 1 &  (u, u) & 9 & 4   \\
  & 4PT$_{1uv}f_{1,2}$ & 3 & 0 & 1 & 1 &  (u, v) & 10 & 16    \\
  & \red{3PT$_{s00}f_{1,2}$} & 3 & 0 & 0 & s &  (0, 0) & 9 & 1   \\
  & 4PT$_{suu}f_{1,2}$ & 3 & 0 & 1 & s &  (u, u) & 10 & 18   \\
  & \red{4PT$_{suv}f_{1,2}$} & 3 & 1 & 0 & s &  (u, v) & 11 & 4   \\
  \bottomrule
  \end{tabular} 
  }
\end{center}
\vspace{-0.15in}
  \caption{Different combinations of using 3D-3D, 3D-2D and 2D-2D point correspondences for focal length problems. The \blue{blue} and \red{red} markers represent the solvers used in our experiments.
  %, which are designed for scale-invariant and affine-invariant depth, respectively. 
  Note that subscripts $suv$ in names of the solvers represent different configurations of known/unknown/same/different scales and shifts in the depth maps.
  $^*$Relative depth was used, where one partial 3D-3D correspondence provides only 2 constraints.}
  \label{tab:solvers}   
   \vspace{-5mm}
\end{table}
%\section{Relative Pose Estimation}

\subsection{Calibrated Case}\label{sec:calib}
In this section, we first discuss the solver for calibrated case, where the intrinsic parameters of cameras are known, \ie, %the calibration matrices 
$\M K_1$ and $\M K_2$ are known.
%from calibration or EXIF. 
We use $\tilde{\M q}_i = \M K_2^{-1} \M q_{i}$, $\tilde{\M p}_i = \M K_1^{-1} \M p_{i}$ to represent the normalized image points.

% \subsection{Depth up to Scale and Shift}

Given 3 point correspondences and their monocular depths, substituting $\tilde{\M q}_i, \tilde{\M p}_i,\ i=1,2,3$ into~\eqref{eq:04} we have nine equations in nine unknowns
\begin{equation}
 s(\beta_{i} + v)\tilde{\M q}_i  =(\alpha_{i} +u) \M R \tilde{\M p}_i  +{\M t}.\label{eq:05} 
\end{equation}
By eliminating the translation from the above equations, \ie, by subtracting pairs of equations, we have
\begin{equation}
\resizebox{0.95\hsize}{!}{%
$
\begin{aligned}
% s(\beta_1+v)\tilde{\M q}_1 - s(\beta_2+v)\tilde{\M q}_2   =\M R ((\alpha_1+u)\tilde{\M p}_1 - (\alpha_2+u)\tilde{\M p}_2), \\
% s(\beta_1+v)\tilde{\M q}_1 - s(\beta_3+v)\tilde{\M q}_3   =\M R ((\alpha_1+u)\tilde{\M p}_1 - (\alpha_3+u)\tilde{\M p}_3), \\
% s(\beta_2+v)\tilde{\M q}_2 - s(\beta_3+v)\tilde{\M q}_3   =\M R ((\alpha_2+u)\tilde{\M p}_2 - (\alpha_3+u)\tilde{\M p}_3). 
s(\beta_i+v)\tilde{\M q}_i - s(\beta_j+v)\tilde{\M q}_j   =\M R ((\alpha_i+u)\tilde{\M p}_i - (\alpha_j+u)\tilde{\M p}_j), \end{aligned}$}\nonumber
\end{equation}
 for $(i,j)=(1,2),(1,3),(2,3)$.
 
Since applying the rotation matrix on a vector preserves the length of this vector, we have the following constraints
\begin{equation}
\resizebox{0.95\hsize}{!}{%
$
\begin{aligned}
%\|s(\beta_1+v)\tilde{\M q}_1 - s(\beta_2+v)\tilde{\M q}_2 \|  =\| (\alpha_1+u)\tilde{\M p}_1 - (\alpha_2+u)\tilde{\M p}_2 \|, \\
%\|s(\beta_1+v)\tilde{\M q}_1 - s(\beta_3+v)\tilde{\M q}_3 \|  =\|(\alpha_1+u)\tilde{\M p}_1 - (\alpha_3+u)\tilde{\M p}_3 \|, \\
%\|s(\beta_2+v)\tilde{\M q}_2 - s(\beta_3+v)\tilde{\M q}_3 \|  =\|(\alpha_2+u)\tilde{\M p}_2 - (\alpha_3+u)\tilde{\M p}_3 \|,
\|s(\beta_i+v)\tilde{\M q}_i - s(\beta_j+v)\tilde{\M q}_j \|  =\| (\alpha_i+u)\tilde{\M p}_i - (\alpha_j+u)\tilde{\M p}_j \|, \\
\end{aligned}$}\nonumber
\end{equation}
 for $(i,j)=(1,2),(1,3),(2,3)$. In this case, the rotation is eliminated and we obtain three equations in three unknowns. Since the equations only contain $s^2$, we can let $c=s^2$ to reduce the degrees and eliminate the symmetries. In short, if $s$ is a solution, then $-s$ is also a solution. However, $s$ should be positive since it presents the relative scale of two depths. thus a solution to $c = s^2$ gives us only one geometrically feasible solution for $s$.

The three above equations can be written in a matrix form $\M M\ [cv^2,cv,c, u^2, u, 1]^\top =0,$
%\begin{equation}
%    \M M\ [cv^2,cv,c, u^2, u, 1]^\top =0,
%    %\ i=1,2,3,
%    \label{eq:06}
%\end{equation}
%where $\M m_i$ is a $1\times 6$ coefficient vector.
where $\M M$ is a $3 \times 6$ coefficient matrix. After Gauss-Jordan (G-J) elimination of this matrix, the three monomials $\{cv^2,cv,c\}$ can be expressed as quadratic functions of $u$,~\ie, $cv^2=g_1(u)$, $cv=g_2(u)$, $c=g_3(u)$, where $g_1,g_2,g_3$ are polynomials in $u$ of degree 2. Since $(cv)^2 = (c)(cv^2)$, we have $g_2^2=g_1 g_3$, resulting in a quartic equation in $u$. This equation can be solved in closed form. We denote the final solver as \textbf{3PT$_{suv}$}. 
%If we have scale-invariant depth, we can use the P3P solver to find the rotation and translation since using a single-side depth would not be affected by the scale. In table~\ref{tab:solvers}, we show details for different solvers using different number of correspondences. 
Note that the inverse depth model results in a more complex solver without improving performance in our experiments (details provided in the SM). Therefore, we do not discuss the inverse depth model in the main paper.

\subsection{Focal Length Problems}

% \begin{equation}
%  s_2(\beta_{i} + v)\M K_2^{-1} \M q_{i} =s_1(\alpha_{i} + u) \M R \M K_1^{-1} \M p_{i}  +\M T, \label{eq:12}
% \end{equation}

In many practical scenarios, the intrinsic parameters of the cameras may not be available. However, it is often reasonable to assume that modern cameras have square-shaped pixels, and the principal point coincides with the image center~\cite{hartley2012efficient}. This is a widely used assumption in many camera geometry solvers, where the only unknown intrinsic parameters are focal lengths, 
%and the calibration matrices have the form 
\ie, $\M K_i = {\rm diag}(f_i,f_i,1)$.

\begin{table}[t]
	\begin{center}
    \resizebox{0.7\linewidth}{!}
        { 
    \begin{tabular}{lccc}
        \toprule
         Solver  &   G-J & Eigen &  Time ($\mu$s) \\ 
        \midrule
        {3PT$_{suv}$}(Ours) & $3\times 6$ & Closed-form & 1.46 \\
    3PT$_{suv}$(M)  & $12\times16$   & $4\times4$ & 4.45 \\
    \midrule
       {4PT$_{suv}f$}(Ours) & $24\times 32$ & $8\times 8$ & 12.5 \\
    4PT$_{suv}f$(M)  & $36\times44$   & $8\times 8$ & 23.6 \\
        \midrule
        {4PT$_{suv}f_{1,2}$}(Ours) & $20\times 24$ & $4\times4$ & 6.45 \\
    4PT$_{suv}f_{1,2}$(M)  & $40\times44$   & $4\times4$ & 20.2 \\
        \bottomrule
    \end{tabular}
        }
	\end{center}
    \vspace{-5mm}
\caption{Efficiency Comparison between the proposed affine-invariant solvers and those from Madpose~\cite{yu2025relative}, denoted as (M).}
	\label{table:eff}
    \vspace{-4mm}
\end{table}

\vspace{1mm}
\noindent{\bf Shared Unknown Focal Length}
\vspace{1mm}

First, let us assume that the two cameras have a shared unknown focal length, \ie, $\M K_1 =\M K_2 ={\rm diag}(f,f,1)$, which is a common scenario, \eg, when estimating the motion of a single uncalibrated camera. This is a 10-DOF problem w.r.t. $\{s,u,v,\M R, {\M t},f\}$, and we need at least three 3D-3D point correspondences and one 2D-2D point correspondence, \ie four 2D-2D correspondences from which three have monocular depths estimated in both images and thus can be considered as 3D-3D correspondences. 
%Note that, here the 3D-3D point correspondences are formulated with the focal length parameter. 
Note that, here the 3D-3D point correspondences depend on the unknown focal length parameter.

In this case, we may obtain the following system of 6 equations in 6 unknowns $\{s,u,v,f,\eta_4,\lambda_4\}$
\begin{equation}
\resizebox{0.99\hsize}{!}{%
$
\begin{aligned}
% \|s\M K_2^{-1}((\beta_1+v){\M q}_1 - s(\beta_2+v){\M q}_2) \|  &=\|\M K_1^{-1} ((\alpha_1+u){\M p}_1 - (\alpha_2+u){\M p}_2) \|, \\
% \|s\M K_2^{-1}((\beta_1+v){\M q}_1 - s(\beta_3+v){\M q}_3) \|  &=\|\M K_1^{-1} ((\alpha_1+u){\M p}_1 - (\alpha_3+u){\M p}_3) \|, \\
% \|s\M K_2^{-1}((\beta_2+v){\M q}_2 - s(\beta_3+v){\M q}_3) \|  &=\|\M K_1^{-1} ((\alpha_2+u){\M p}_2 - (\alpha_3+u){\M p}_3) \|, \\
% \|s\M K_2^{-1}((\beta_1+v){\M q}_1 - s\lambda_4{\M q}_j) \|  &=\|\M K_1^{-1} ((\alpha_1+u){\M p}_1 - \eta_4{\M p}_j) \|, \\
% \|s\M K_2^{-1}((\beta_2+v){\M q}_2 - s\lambda_4{\M q}_4) \|  &=\|\M K_1^{-1} ((\alpha_2+u){\M p}_2 - \eta_4{\M p}_4) \|, \\
% \|s\M K_2^{-1}((\beta_3+v){\M q}_3 - s\lambda_4{\M q}_4) \|  &=\|\M K_1^{-1} ((\alpha_3+u){\M p}_3 - \eta_4{\M p}_4) \|. 
\|s\M K_2^{-1}((\beta_i+v){\M q}_i - s(\beta_j+v){\M q}_j) \|  &=\|\M K_1^{-1} ((\alpha_i+u){\M p}_i - (\alpha_j+u){\M p}_j) \|, \\
\|s\M K_2^{-1}((\beta_k+v){\M q}_k - \lambda_4{\M q}_4) \|  &=\|\M K_1^{-1} ((\alpha_k+u){\M p}_k - \eta_4{\M p}_4) \|, 
\end{aligned}$}\nonumber
\end{equation}
 for $(i,j)=(1,2),(1,3),(2,3)$ and $k=1,2,3$.
To solve such a system, we need to perform G-J elimination of a large matrix. Moreover, the system has up to 28 real solutions. 
Thus, its solver is not practically relevant.
%relevant for practical applications.

To find a more efficient and practical solution, we use four 2D-2D point correspondences with their four depths in both images.
In this case, we obtain the following 6 equations in 4 unknowns $\{s,u,v,f\}$
\begin{equation}
\resizebox{0.99\hsize}{!}{%
$
\begin{aligned}
\|s\M K_2^{-1}((\beta_i+v){\M q}_i - (\beta_j+v){\M q}_j) \|  &=\|\M K_1^{-1} ((\alpha_i+u){\M p}_i - (\alpha_j+u){\M p}_j) \|, 
\end{aligned}$}\nonumber
\end{equation}
where $i,j = 1,2,3,4,\ i\neq j$. In general, it is an over-constrained system.
One way to solve it is to use four of these six equations in four unknowns. The system of four polynomials can be solved using the Gr\"{o}bner basis method~\cite{larsson2017efficient}, where the final solver performs G-J elimination of a $24 \times 32$ matrix and extracts up to 8 real solutions from the eigenvalues and eigenvectors of an $8 \times 8$ matrix. We denote this solver as \textbf{4PT$_{suv}f$}.
Alternatively, by using all six equations, we can derive a 
significantly 
smaller solver 
%for this problem 
with up to two real solutions. This solver only needs to perform the G-J elimination of a $6 \times 8$ matrix. However, it is more sensitive to noise. Further details on this simplified focal length solver are provided in the SM. 

With scale invariant depth, we have an 8-DOF problem w.r.t. $\{s, \M R, \M t, f\}$. We need at least two 3D-3D point correspondences and one 3D-2D point correspondence. The problem can be solved similarly as the calibrated case shown in Sec.~\ref{sec:calib}. Due to the space limits, the details are shown in the SM, where we show a general solution to all problem configurations using three point correspondences.

\vspace{1mm}
\noindent{\bf Different and Unknown Focal Lengths}
\vspace{1mm}

Finally, we consider the case where $\M K_i = {\rm diag}(f_i,f_i,1), \; i=1,2$, with $f_1 \neq f_2$. This is a 11-DOF problem w.r.t. $\{s,u,v,\M R, {\M t},f_1,f_2\}$. Thus, we need at least three 3D-3D point correspondences and one 3D-2D point correspondence, \ie a correspondence where we have monocular depth only in one image. Similarly to the equal and unknown focal length case, we can formulate this problem using a system of 6 equations in 6 unknowns with up to 14 real solutions and solved it using 
%a Gr\"{o}bner basis method
~\cite{larsson2017efficient}.

%To develop an efficient solution, we use four 3D-3D point correspondences, yielding six equations with five unknowns: ${s, u, v, f_1, f_2}$.
However, a more efficient solver can be obtained using four 3D-3D point correspondences, which yields six equations in five unknowns: ${s, u, v, f_1, f_2}$.
This is an over-constrained system that can be again solved by using only five of these six equations. The five equations can be rewritten as  $\M M\ [1,c,f_1,uf_1,u^2f_1,cf_2,cvf_2,cv^2f_2]^\top =0$,
% \begin{equation}
%     %\M m_i\ [1,c,f_1,uf_1,u^2f_1,cf_2,cvf_2,cv^2f_2]^\top =0,
%     \M M\ [1,c,f_1,uf_1,u^2f_1,cf_2,cvf_2,cv^2f_2]^\top =0,
%     \label{eq:07}
% \end{equation}
where %$\M m_{i,i=1,2,...,5}$ is a $1\times 8$ 
%coefficient vector.
$\M M$ is a $5\times 8$. 
%We notice that 
Here,
$f_2$ always appears together with $c$, thus, we let $cf_2 = \tilde{f}_2$ to simplify the polynomials. In this case, we obtain a solver that performs the 
%Gauss-Jordan
G-J
elimination of a $20 \times 24$ matrix and results in up to four solutions. We denote this solver as \textbf{4PT$_{suv}f_{1,2}$}.
%Similarly,
Alternatively, we can use all six equations to derive a smaller solver, which is faster, but more sensitive to noise (see SM for more details).
%We refer the readers to SM for more details.

With scale-invariant depth, we have a 9-DOF problem w.r.t. $\{s, \M R, \M t, f_1, f_2\}$. We need at least three 3D-3D point correspondences. This problem can be solved similarly as the calibrated case shown in Sec.~\ref{sec:calib}. Due to space limits, the details on this \textbf{3PT$_{s00}f_{1,2}$} solver are shown in the SM.

%In table~\ref{tab:solvers}, we show the details for different focal length solvers with different depth properties. The \blue{blue} (assuming scale-invariant depth) and \red{red} (assuming affine-invariant depth) are more practical and will be evaluated in our experiments. Since~\cite{yu2025relative} considers only affine-invariant depth, we compare the efficiency of our solvers with those from~\cite{yu2025relative} under affine-invariant depth in Table~\ref{table:eff}.

% \begin{table}[t]
% 	\begin{center}
%     \resizebox{0.7\linewidth}{!}
%         { 
%     \begin{tabular}{lccc}
%         \toprule
%          Solver  &   G-J & Eigen &  Time ($\mu$s) \\ 
%         \midrule
%         {3PT$_{suv}$}(Ours) & $3\times 6$ & Closed-form & 1.46 \\
%     3PT$_{suv}$(M)  & $12\times16$   & $4\times4$ & 4.45 \\
%     \midrule
%        {4PT$_{suv}f$}(Ours) & $24\times 32$ & $8\times 8$ & 12.5 \\
%     4PT$_{suv}f$(M)  & $36\times44$   & $8\times 8$ & 23.6 \\
%         \midrule
%         {4PT$_{suv}f_{1,2}$}(Ours) & $20\times 24$ & $4\times4$ & 6.45 \\
%     4PT$_{suv}f_{1,2}$(M)  & $40\times44$   & $4\times4$ & 20.2 \\
%         \bottomrule
%     \end{tabular}
%         }
% 	\end{center}
%     \vspace{-5mm}
% \caption{Efficiency Comparison between the proposed affine-invariant solvers and those from Madpose~\cite{yu2025relative}, denoted as (M).}
% 	\label{table:eff}
%     \vspace{-3mm}
% \end{table}
\section{Experiments}

In this section, we present a comprehensive comparison of different depth-aware and point-based solvers for calibrated, shared focal length and different focal length cases, under different conditions, \ie different sources of depth information, different feature detectors and matchers, different type of datasets (interior/exterior) and different optimization schemes inside RANSAC. 

% \begin{table}[htbp]
% % \input{tables/eth_calib}
% \input{tables/eth_calib_featcol}
% \caption{Comparison of different methods on the \ETH dataset~\cite{schops2017multi} for the calibrated case. Opt.: S - PoseLib~\cite{poselib} implementation using Sampson error, H - hybrid RANSAC from~\cite{yu2025relative}.}
% \label{tab:real_calib}
% \end{table}

\noindent\textbf{Solvers.} (1) For the calibrated relative pose estimation, we compare our 3PT$_{suv}$(ours) solver with the solver 3PT$_{suv}$(M) from~\cite{yu2025relative}, the Rel3PT solver~\cite{Astermark2024}, the P3P solver~\cite{ding2023revisiting}, and the 5PT point-based solver~\cite{nister2004efficient}. (2) For the unknown shared focal length case, we compared the proposed 3PT$_{s00}f$(ours) and 4PT$_{suv}f$(ours) solvers with the 4PT$_{suv}f$(M) solver from~\cite{yu2025relative}, the 3p3d solver~\cite{dingfundamental} and the 6PT point-based solver~\cite{larsson2017efficient}. (3) For the different focal length case, we compare our 3PT$_{s00}f_{1,2}$(ours) and 4PT$_{suv}f_{1,2}$(ours) solvers with the 4PT$_{suv}f_{1,2}$(M) solver from~\cite{yu2025relative}, the 4p4d solver~\cite{dingfundamental} and the 
%point-based 
7PT solver~\cite{hartley2003multiple}.

% We evaluated the performance of the proposed solvers, namely, the 3PT$_{suv}$(GB) 3PT$_{suv}$(Eigen) solvers for the calibrated case, the 4PT$_{suv}f$(GB) and 4PT$_{suv}f$(Eigen) solvers for unknown equal focal lengths, and the 4PT$_{suv}f_{1,2}$(GB) and 4PT$_{suv}f_{1,2}$(Eigen) solvers for varying focal lengths—on both synthetic data and real-world images.   

\noindent\textbf{Datasets.} In order to test the proposed solvers on real-world data, we choose the \Phototourism ~\cite{Jin2020}, the \ETH~\cite{schops2017multi} and the \ScanNet~\cite{dai2017scannet} datasets. 
% They cover three situations: unordered images for Structure-from-Motion, sequential images in an outdoor environment, and sequential images in an indoor environment. 
% Fig.~\ref{fig:example} shows example images and their corresponding disparity maps. 
The images of the \Phototourism dataset were collected from multiple cameras obtained at different times, from different viewpoints, and with occlusions. It is a challenging outdoor dataset that is commonly used as a benchmark dataset~\cite{Jin2020} 
%and can be used to evaluate the performance of methods in a wide range of situations.
for camera geometry methods.
%in a wide range of situations.
We used five test scenes with $24,750$ image pairs from this dataset. 
%including images, ground-truth poses, and sparse depth maps. 
The \ETH multi-view stereo and 3D reconstruction benchmark~\cite{schops2017multi} includes a range of indoor and outdoor scenes. Ground truth geometry was acquired using a high-precision laser scanner. Images were captured with a DSLR camera and a synchronized multi-camera rig featuring cameras with varying fields of view. In total, $4,144$ image pairs were used from the \ETH dataset, and we used triangulation to obtain the ground truth depths. The \ScanNet dataset is an indoor RGB-D video dataset containing 2.5 million views in more than 1500 scans, annotated with 3D camera poses and surface reconstructions.
%, and instance-level semantic segmentations. 
We followed the evaluation setup in~\cite{sarlin2020superglue}, and used 1500 pairs from \ScanNet for testing. 

\noindent\textbf{Feature Detection and Matching.} We used three different image features, the first is the popular sparse SuperPoint (SP) features~\cite{detone2018superpoint}, widely used in SfM and localization pipelines~\cite{sarlin2019coarse}. For SuperPoint matching, we employ LightGlue (LG)~\cite{lindenberger2023lightglue}, a state-of-the-art deep learning-based method that demonstrates significant improvements over traditional matching methods. The second is RoMa~\cite{edstedt2024roma}, a state-of-the-art dense matcher. The last is Mast3r~\cite{leroy2024grounding}, where feature matching is considered as a 3D task with DUSt3R~\cite{wang2024dust3r}, a recent and powerful 3D reconstruction framework based on Transformers.

% \noindent\textbf{Depth Estimation.} We used 11 different monocular depth estimation methods to obtain the depth data, including MiDas~\cite{birkl2023midas}, DPT~\cite{ranftl2021vision}, ZoeDepth~\cite{bhat2023zoedepth}, DA V1 (Depth Anything V1)~\cite{yang2024depth}, DA V2 (Depth Anything V2)~\cite{yang2024depthv2}, Depth Pro~\cite{bochkovskii2024depth}, UniDepth~\cite{piccinelli2024unidepth}, Metric 3d V2~\cite{hu2024metric3d}, Marigold +FT (Fine tuning)~\cite{martingarcia2024diffusione2eft},  MoGe~\cite{wang2024moge}, and Mast3r~\cite{leroy2024grounding}. 
% While using depth maps does not add computational overhead, incorporating learning-based methods does introduce additional costs. However, in some applications, depth estimation methods serve multiple tasks, so this computational cost is not solely attributable to the pose estimation method.

\noindent\textbf{Depth Estimation.} We use five state-of-the-art MDE methods to obtain depth data, including MiDas~\cite{birkl2023midas}, DA v2 (Depth Anything v2)~\cite{yang2024depthv2}, UniDepth~\cite{piccinelli2024unidepth},  MoGe~\cite{wang2024moge}, and Mast3r~\cite{leroy2024grounding}. These MDE methods represent scale/affine-invariant~\cite{birkl2023midas,yang2024depthv2,wang2024moge,leroy2024grounding} as well as metric depths~\cite{piccinelli2024unidepth}. 
%Although using depth maps does not add computational overhead in pose estimation, obtaining these depth maps using learning-based methods potentially introduces additional costs. This is usually a few milliseconds per image~\cite{yu2025relative}. However, in many applications, MDE methods are used for multiple tasks. Thus, the computational overhead is not solely attributable to the pose estimation method. 

% Due to the space limitation, 

\noindent\textbf{Robust Estimation Frameworks.}
We evaluated all solvers using the LO-RANSAC framework~\cite{chum2003locally} within PoseLib~\cite{poselib} with Sampson error for LO and scoring with an error threshold set at 2 px and fixed 1000 RANSAC iterations.
% For example, we construct the essential matrix based on the rotation and translation from the P3P solver, then Sampson error related to this essential matrix is used for scoring and LO. 
We provide additional experiments using the reprojection error in the SM. We also evaluated selected solvers using the Hybrid RANSAC framework presented in~\cite{yu2025relative} with 2 px Sampson error and 16 px reprojection error thresholds with equal weight for both types of error and fixed 1000 iterations.  For Mast3r~\cite{leroy2024grounding}, we use their non-linear optimization 
%strategy 
with 500 iterations for coarse refinement and 200 iterations for fine refinement. 
%With a sufficient number of iterations, we achieved better results using Mast3r compared to the results reported for Mast3r in~\cite{yu2025relative}.
We observe that Mast3r performs better in our evaluation compared to the results reported in~\cite{yu2025relative}. 
We attribute this potentially to using more refinement iterations, which in our experience improves performance.

\noindent\textbf{Evaluation Metrics.}
%To compare the methods 
In experiments, we report the median pose error $\epsilon$ calculated as the maximum of rotation and translation errors in degrees. We also report the mean average accuracy (mAA)~\cite{Jin2020} with a  threshold of 10 degrees. For the focal length solvers we use $f_{err} = |f_{est} - f_{gt}|/f_{gt}$ as the focal length error for one camera. When two cameras with different focal lengths are considered, we use the geometric mean of their errors. We report the median focal length errors denoted as $\epsilon_f$. We also report the mean average accuracy for focal lengths~\cite{kocur2024robust} at the 10\% error threshold, denoted as mAA$_f$. 
%We also report run-times in milliseconds denoted as $\tau$. 
Run-times $\tau$ are reported in milliseconds.
Run-times for all methods except Mast3r are reported per one CPU of Intel Xeon Gold 6338. The run-times of Mast3r are reported %based 
on the CPU of Intel I7-11700KF with RTX 3090 GPU. %Note that 
The run-times of all the methods do not include the feature matching and depth estimation since they are equal for all depth-based solvers. For a fair comparison, for Mast3r, we only measure the run-times of the non-linear optimization, \ie time without feature matching, and initial depth estimation. For datasets with multiple scenes, we report the mean values across the scenes.

\subsection{Calibrated Relative Pose Estimation}

Table~\ref{tab:real_calib} shows the comparison of different methods on the \ETH dataset. The results for the \Phototourism and \ScanNet datasets are provided in the SM. The results show that using depth in pose estimation results in more accurate poses when good depth estimates (e.g. MoGe~\cite{wang2024moge}, UniDepth~\cite{piccinelli2024unidepth}) are available. In this case, the overall best-performing methods use the hybrid RANSAC proposed in ~\cite{yu2025relative} with our solver 3PT$_{suv}$(ours) slightly outperforming 3PT$_{suv}$(M)~\cite{yu2025relative}. We note that the use of the hybrid RANSAC scheme comes at a significant increase in computational cost over the use of PoseLib~\cite{poselib}. As the authors of~\cite{yu2025relative} point out, their implementation of the hybrid RANSAC scheme, which we use in our experiments, could be 
%significantly 
optimized for computational efficiency. However, even with optimizations the method requires significantly more computation per one iteration than standard RANSAC. 

Based on the results for PoseLib~\cite{poselib}, when good depth is available, P3P~\cite{ding2023revisiting} performs better than solvers which also model shift. When less accurate depth estimates are available, the 5PT performs best especially with good matches (e.g., Mast3r~\cite{leroy2024grounding}, RoMA~\cite{edstedt2024roma}). This opens up potential for future work, which would incorporate the P3P solver into the hybrid RANSAC scheme proposed in~\cite{yu2025relative}.

\begin{table}[htbp]
\resizebox{\linewidth}{!}{
\begin{tabular}{clccccccc}
\toprule
 \multirow{2.5}{*}{{Depth}} &  \multirow{2.5}{*}{{Solver}} & \multirow{2.5}{*}{{Opt.}} & \multicolumn{3}{c}{SP+LG~\cite{detone2018superpoint, lindenberger2023lightglue}}& \multicolumn{3}{c}{RoMA~\cite{edstedt2024roma}}\\
\cmidrule(rl){4-6} \cmidrule(rl){7-9}
& & & $\epsilon(^\circ)\downarrow$ & mAA$\uparrow$ & $\tau (ms)\downarrow$  & $\epsilon(^\circ)\downarrow$ & mAA $\uparrow$ & $\tau (ms)\downarrow$ \\
\midrule
\multirow{ 1 }{*}{ \makecell{-} }
& 5PT~\cite{nister2004efficient} & S & 0.91&87.67&48.14&0.56&91.10&184.36 \\
\hline
\multirow{ 6 }{*}{ \makecell{Real \\ Depth} }
& Rel3PT~\cite{Astermark2024} & S & 0.88&88.21&103.18&0.52&91.38&532.02 \\
& P3P~\cite{ding2023revisiting} & S & 0.83&88.88&\underline{29.68}&0.52&91.33&\underline{141.39} \\
& 3PT$_{suv}$(M)~\cite{yu2025relative} & S & 0.79&88.55&41.81&0.45&91.39&145.59 \\
& 3PT$_{suv}$(\textbf{ours}) & S & 0.80&88.60&\textbf{29.59}&0.47&91.37&\textbf{127.81} \\
& 3PT$_{suv}$(M)~\cite{yu2025relative} & H~\cite{yu2025relative} & \underline{0.52}&\underline{91.39}&549.59&\underline{0.39}&\textbf{92.73}&1505.19 \\
& 3PT$_{suv}$(\textbf{ours}) & H~\cite{yu2025relative} & \textbf{0.52}&\textbf{91.42}&543.48&\textbf{0.39}&\underline{92.72}&1490.93 \\
\hline
\multirow{ 6 }{*}{ \makecell{MiDas \\ \cite{birkl2023midas}} }
& Rel3PT~\cite{Astermark2024} & S & 4.81&71.25&36.34&3.23&82.22&149.63 \\
& P3P~\cite{ding2023revisiting} & S & 0.94&86.16&\underline{22.97}&0.60&\textbf{90.80}&91.36 \\
& 3PT$_{suv}$(M)~\cite{yu2025relative} & S & \textbf{0.88}&\underline{87.34}&31.11&\textbf{0.58}&\underline{90.77}&\underline{79.17} \\
& 3PT$_{suv}$(\textbf{ours}) & S & \underline{0.88}&\textbf{87.39}&\textbf{20.40}&\underline{0.59}&90.76&\textbf{67.13} \\
& 3PT$_{suv}$(M)~\cite{yu2025relative} & H~\cite{yu2025relative} & 1.08&85.38&685.32&0.67&90.49&1605.43 \\
& 3PT$_{suv}$(\textbf{ours}) & H~\cite{yu2025relative} & 1.07&85.45&683.39&0.67&90.50&1590.71 \\
\hline
\multirow{ 6 }{*}{ \makecell{DA v2 \\ \cite{yang2024depthv2}} }
& Rel3PT~\cite{Astermark2024} & S & 5.57&68.46&35.74&3.90&80.56&145.85 \\
& P3P~\cite{ding2023revisiting} & S & \underline{0.90}&86.25&\underline{23.26}&0.72&90.61&93.99 \\
& 3PT$_{suv}$(M)~\cite{yu2025relative} & S & \textbf{0.90}&\textbf{87.19}&32.50&0.56&\underline{90.99}&\underline{88.60} \\
& 3PT$_{suv}$(\textbf{ours}) & S & 0.91&\underline{87.07}&\textbf{21.68}&0.56&\textbf{91.01}&\textbf{75.87} \\
& 3PT$_{suv}$(M)~\cite{yu2025relative} & H~\cite{yu2025relative} & 0.97&85.52&605.63&\underline{0.54}&90.62&1493.75 \\
& 3PT$_{suv}$(\textbf{ours}) & H~\cite{yu2025relative} & 0.98&85.56&593.95&\textbf{0.54}&90.62&1477.73 \\
\hline
\multirow{ 6 }{*}{ \makecell{MoGe \\ \cite{wang2024moge}} }
& Rel3PT~\cite{Astermark2024} & S & 4.74&72.08&42.19&2.74&82.04&170.29 \\
& P3P~\cite{ding2023revisiting} & S & 0.91&87.67&\underline{25.72}&0.54&91.16&111.74 \\
& 3PT$_{suv}$(M)~\cite{yu2025relative} & S & 0.89&87.71&33.45&0.53&91.04&\underline{98.20} \\
& 3PT$_{suv}$(\textbf{ours}) & S & 0.89&87.67&\textbf{22.41}&0.54&91.05&\textbf{84.24} \\
& 3PT$_{suv}$(M)~\cite{yu2025relative} & H~\cite{yu2025relative} & \underline{0.86}&\textbf{88.26}&566.16&\underline{0.50}&\underline{91.17}&1414.96 \\
& 3PT$_{suv}$(\textbf{ours}) & H~\cite{yu2025relative} & \textbf{0.85}&\underline{88.24}&554.79&\textbf{0.49}&\textbf{91.23}&1401.61 \\
\hline
\multirow{ 6 }{*}{ \makecell{UniDepth \\ \cite{piccinelli2024unidepth}} }
& Rel3PT~\cite{Astermark2024} & S & 1.36&78.82&49.70&0.70&88.25&207.86 \\
& P3P~\cite{ding2023revisiting} & S & 0.88&88.00&\underline{25.93}&0.56&91.11&112.65 \\
& 3PT$_{suv}$(M)~\cite{yu2025relative} & S & 0.94&87.42&33.90&0.55&91.04&\underline{97.85} \\
& 3PT$_{suv}$(\textbf{ours}) & S & 0.95&87.49&\textbf{22.58}&0.55&91.01&\textbf{83.72} \\
& 3PT$_{suv}$(M)~\cite{yu2025relative} & H~\cite{yu2025relative} & \textbf{0.86}&\underline{88.03}&558.61&\underline{0.53}&\underline{91.33}&1402.46 \\
& 3PT$_{suv}$(\textbf{ours}) & H~\cite{yu2025relative} & \underline{0.86}&\textbf{88.03}&550.74&\textbf{0.53}&\textbf{91.33}&1392.71 \\
\hline \\
% \end{tabular}
\multicolumn{9}{c}{
\begin{tabular}{clcccc}
\hline
\multirow{2.5}{*}{{Depth}} &  \multirow{2.5}{*}{{Solver}} & \multirow{2.5}{*}{{Opt.}} & \multicolumn{3}{c}{Mast3r~\cite{leroy2024grounding}} \\ \cmidrule{4-6}
&&& $\epsilon(^\circ)\downarrow$ & mAA $\uparrow$ & $\tau (ms)\downarrow$ \\ \cmidrule{1-6}
\multirow{ 1 }{*}{ \makecell{-} }
& 5PT~\cite{nister2004efficient} & S & 0.66&90.29&126.77 \\
\cmidrule{1-6}
\multirow{ 6 }{*}{Mast3r~\cite{leroy2024grounding}}
& Rel3PT~\cite{Astermark2024} & S & 0.67&\underline{90.30}&104.46 \\
& P3P~\cite{ding2023revisiting} & S & \underline{0.67}&90.24&56.52 \\
& 3PT$_{suv}$(M)~\cite{yu2025relative} & S & 0.67&90.20&\underline{42.49} \\
& 3PT$_{suv}$(\textbf{ours}) & S & \textbf{0.67}&\textbf{90.35}&\textbf{30.90} \\
& 3PT$_{suv}$(M)~\cite{yu2025relative} & H~\cite{yu2025relative} & 0.92&87.96&2647.02 \\
& 3PT$_{suv}$(\textbf{ours}) & H~\cite{yu2025relative} & 0.92&87.98&2635.27 \\
\cmidrule{1-6}
\end{tabular}}
\end{tabular}}

\caption{Comparison of different methods on the \ETH dataset~\cite{schops2017multi} for the calibrated case. Opt.: S - PoseLib~\cite{poselib} implementation using Sampson error, H - hybrid RANSAC from~\cite{yu2025relative}.}
\label{tab:real_calib}
\end{table}

%With the scale and shift formulation, our solver gives more stable results for different monocular depths. With very accurate depth, the P3P algorithm performs slightly better, but the P3P algorithm is more sensitive to noise in the depth data. Our solver and the P3P solver using depth from~\cite{wang2024moge,piccinelli2024unidepth} perform slightly better than the 5PT algorithm on \Phototourism. Here, we skip the inverse depth solver 3PT$_{suv}$(inverse) since it is slower and does not give better results than the 3PT$_{suv}$ solver. Results for 3PT$_{suv}$(inverse) are shown in the SM. 

\subsection{Shared Unknown Focal Length}
Table~\ref{tab:real_equal} shows the comparison of the methods on the \ETH dataset~\cite{schops2017multi} for two cameras with a shared unknown focal length. When a good depth estimate is available, the best results are obtained utilizing the hybrid RANSAC scheme from~\cite{yu2025relative}. We show that our solver 3PT$_{s00}f$ outperforms 4PT$_{suv}f$(M)~\cite{yu2025relative} for many combinations of MDEs and matches despite not modeling shift. Our 3PT$_{s00}f$ solver also performs best when comparing solvers incorporated within PoseLib~\cite{poselib} and used in combination with MoGe~\cite{wang2024moge}, UniDepth~\cite{piccinelli2024unidepth} and Mast3r~\cite{leroy2024grounding}. For DA v2~\cite{yang2024depthv2} and MiDas~\cite{birkl2023midas} modeling shift seems to be beneficial, and we show that our 4PT$_{suv}f$ solver performs on par with 4PT$_{suv}f$(M) while being 2x faster. Our observation that modeling an unknown shift is not always necessary is in contradiction to the observation made in~\cite{yu2025relative}. The reason is that~\cite{yu2025relative}, have not tested a scale-invariant 3PT solver
%that uses full depth information 
and have not incorporated other depth-aware solvers into the hybrid ransac scheme. We also note that despite performing expensive non-linear optimization to obtain the poses Mast3r~\cite{leroy2024grounding} performs worse than approaches based on RANSAC when either good depth (UniDepth~\cite{piccinelli2024unidepth}, MoGe~\cite{wang2024moge}) or matches (RoMA~\cite{edstedt2024roma}) are available.

\begin{figure}
    \setlength{\tabcolsep}{0pt} 
    \centering
    % \begin{tabular}
    % {m{0.68\linewidth}m{0.28\linewidth}}
    
    \resizebox{\linewidth}{!}{    
    \begin{tikzpicture} 
        \begin{axis}[%
        hide axis, xmin=0,xmax=0,ymin=0,ymax=0,
        legend style={draw=white!15!white, 
        line width = 1pt,
        legend cell align=left,
        legend  columns =5, % comment for column display
        /tikz/every even column/.append style={column sep=0.05cm},
        font=\scriptsize
        },
        legend image post style={xscale=1}
        ]

        \addlegendimage{white}\addlegendentry{Solvers:}
        \addlegendimage{Seaborn1}        \addlegendentry{3PT$_{s00}f$(\textbf{ours})};
        % \addlegendimage{Seaborn2}        \addlegendentry{3PT$_{suv}f$(\textbf{ours})};        
        \addlegendimage{Seaborn3}
        \addlegendentry{3PT$_{suv}f$(M)~\cite{yu2025relative}};

        \addlegendimage{Seaborn4}        \addlegendentry{3p3d~\cite{dingfundamental}};        
        \addlegendimage{Seaborn6}
        \addlegendentry{6PT~\cite{hartley2012efficient}};
        \addlegendimage{white}\addlegendentry{Estimators:}        \addlegendimage{black!30}\addlegendentry{PoseLib~\cite{poselib}}
        \addlegendimage{black!30,dash pattern=on 2pt off 1pt on 2pt off 1pt}
        \addlegendentry{Hybrid~\cite{yu2025relative}};
        \addlegendimage{Seaborn5,dash pattern=on 1pt off 0.5pt on 1pt off 0.5pt}        \addlegendentry{Mast3r~\cite{leroy2024grounding}};
        \end{axis}
    \end{tikzpicture}}
    \includegraphics[width=0.6\linewidth]{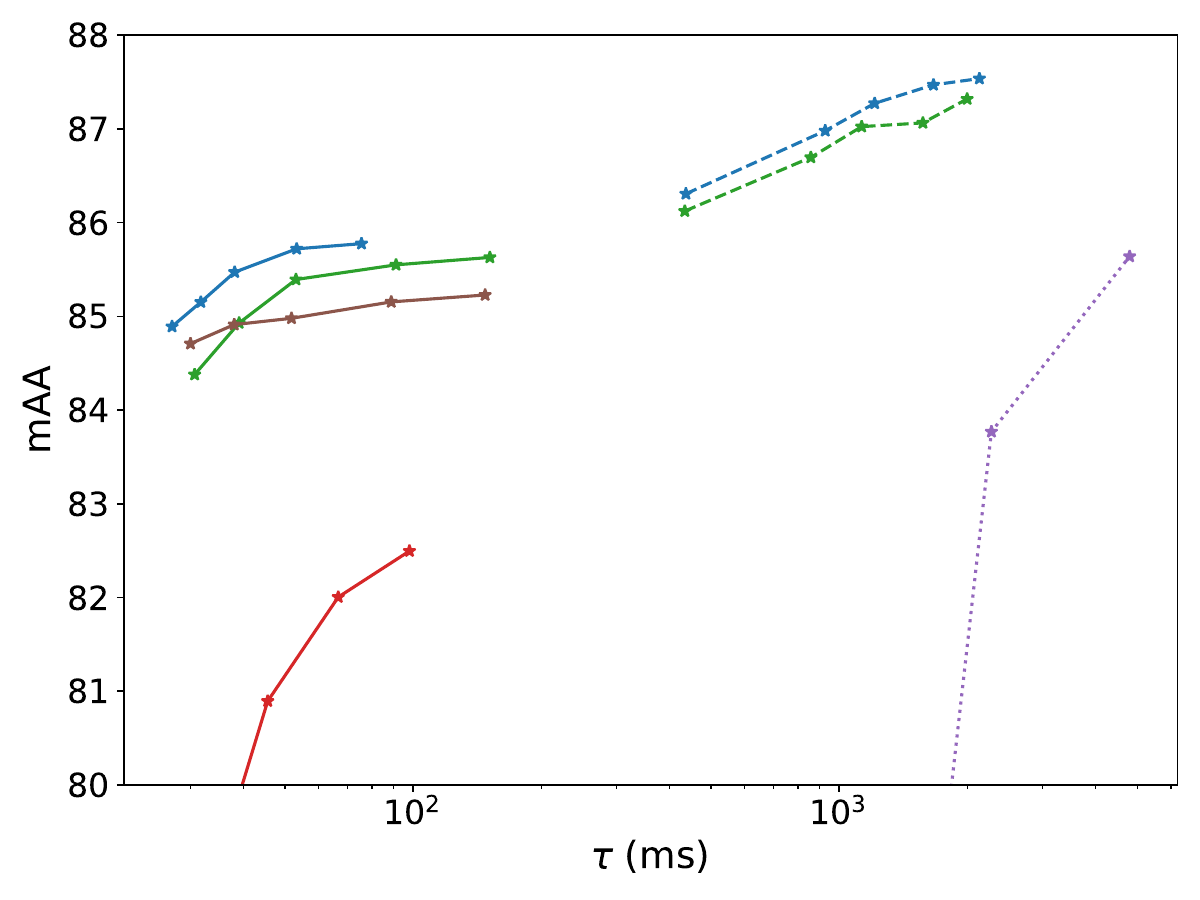}        \vspace{-3mm}
    % \end{tabular}
    \caption{Speed accuracy evaluation for the case of two cameras with shared unknown focal lengths using the RoMA matches~\cite{edstedt2024roma} and UniDepth~\cite{piccinelli2024unidepth} depths on the \ETH dataset~\cite{schops2017multi}. We evaluated mAA and runtimes ($\tau$) for each method running for 50, 100, 200, 500 and 1000 iterations with exception for Mast3r~\cite{leroy2024grounding} which ran with maximum of 500 iterations.}
    \label{fig:shared_speed_acc_roma_eth}
\end{figure}

A comparison of speed-accuracy tradeoff of various methods is shown in Figure~\ref{fig:shared_speed_acc_roma_eth} for the combination of RoMA~\cite{edstedt2024roma} and UniDepth~\cite{piccinelli2024unidepth}. The figure shows that when fast computation is necessary, our solver 3PT$_{s00}f$ implemented in PoseLib~\cite{poselib} performs the best. However, in general, the best accuracy is achieved using the same solver within the hybdrid RANSAC scheme~\cite{yu2025relative}.

\begin{table}[htbp]
    \centering
        \resizebox{\linewidth}{!}{
\begin{tabular}{cclcccccc}
\toprule
% \multirow{2.5}{*}{{Matches}} & \multirow{2.5}{*}{{Depth}} &  \multirow{2.5}{*}{{Solver}} & \multirow{2.5}{*}{{Opt}} & \multicolumn{5}{c}{ETH}\\
% \cmidrule(rl){5-9}
% & & & & $\epsilon(^\circ)\downarrow$ & $\epsilon_f(^\circ)\downarrow$ & mAA$\uparrow$ & mAA$_f\uparrow$ & $\tau (ms)\downarrow$ \\
\multirow{1}{*}{{Matches}} & \multirow{1}{*}{{Depth}} &  \multirow{1}{*}{{Solver}} & \multirow{1}{*}{{Opt.}} & $\epsilon(^\circ)\downarrow$ & $\epsilon_f(^\circ)\downarrow$ & mAA$\uparrow$ & mAA$_f\uparrow$ & $\tau (ms)\downarrow$ \\
\midrule
\multirow{ 31 }{*}{\makecell{ SP+LG \\ \cite{detone2018superpoint, lindenberger2023lightglue} }}
& \multirow{ 1 }{*}{ \makecell{-} }
& 6PT~\cite{larsson2017efficient} & S & 2.45&0.04&75.57&61.52&80.02 \\
\cmidrule(rl){2- 9 }
& \multirow{ 6 }{*}{ \makecell{Real \\ Depth} }
& 3p3d~\cite{dingfundamental} & S & 2.06&0.04&78.00&62.83&\underline{30.70} \\
&
& 4PT$_{suv}f$(M)~\cite{yu2025relative} & S & 1.83&0.03&78.86&63.71&112.34 \\
&
& 4PT$_{suv}f$(\textbf{ours}) & S & 1.72&0.03&78.90&63.56&51.10 \\
&
& 3PT$_{s00}f$(\textbf{ours}) & S & 1.75&0.03&79.17&63.63&\textbf{25.11} \\
&
& 4PT$_{suv}f$(M)~\cite{yu2025relative} & H~\cite{yu2025relative} & \underline{1.07}&\textbf{0.02}&\underline{82.01}&\underline{75.63}&1502.02 \\
&
& 3PT$_{s00}f$(\textbf{ours}) & H~\cite{yu2025relative} & \textbf{1.07}&\underline{0.02}&\textbf{82.19}&\textbf{75.76}&1411.05 \\
\cmidrule(rl){2- 9 }
& \multirow{ 6 }{*}{ \makecell{MiDas \\ \cite{birkl2023midas}} }
& 3p3d~\cite{dingfundamental} & S & 4.08&0.07&61.91&49.69&\underline{25.35} \\
&
& 4PT$_{suv}f$(M)~\cite{yu2025relative} & S & \textbf{2.17}&\textbf{0.04}&\textbf{73.99}&\underline{59.89}&103.86 \\
&
& 4PT$_{suv}f$(\textbf{ours}) & S & \underline{2.26}&\underline{0.04}&\underline{73.53}&59.35&45.74 \\
&
& 3PT$_{s00}f$(\textbf{ours}) & S & 2.26&0.04&73.11&\textbf{60.38}&\textbf{21.36} \\
&
& 4PT$_{suv}f$(M)~\cite{yu2025relative} & H~\cite{yu2025relative} & 2.58&0.05&68.70&56.01&1303.38 \\
&
& 3PT$_{s00}f$(\textbf{ours}) & H~\cite{yu2025relative} & 2.47&0.05&69.04&56.45&1316.09 \\
\cmidrule(rl){2- 9 }
& \multirow{ 6 }{*}{ \makecell{DA v2 \\ \cite{yang2024depthv2}} }
& 3p3d~\cite{dingfundamental} & S & 4.54&0.08&61.60&48.76&\underline{27.01} \\
&
& 4PT$_{suv}f$(M)~\cite{yu2025relative} & S & \underline{2.10}&0.04&\underline{74.60}&\underline{60.77}&104.75 \\
&
& 4PT$_{suv}f$(\textbf{ours}) & S & \textbf{2.02}&\underline{0.04}&\textbf{74.67}&60.41&46.62 \\
&
& 3PT$_{s00}f$(\textbf{ours}) & S & 2.20&\textbf{0.04}&73.66&\textbf{60.93}&\textbf{21.62} \\
&
& 4PT$_{suv}f$(M)~\cite{yu2025relative} & H~\cite{yu2025relative} & 2.34&0.05&70.07&58.00&1190.89 \\
&
& 3PT$_{s00}f$(\textbf{ours}) & H~\cite{yu2025relative} & 2.33&0.05&70.16&58.31&1216.89 \\
\cmidrule(rl){2- 9 }
& \multirow{ 6 }{*}{ \makecell{MoGe \\ \cite{wang2024moge}} }
& 3p3d~\cite{dingfundamental} & S & 3.18&0.06&68.24&54.64&\underline{27.70} \\
&
& 4PT$_{suv}f$(M)~\cite{yu2025relative} & S & 2.10&0.04&76.06&61.81&107.30 \\
&
& 4PT$_{suv}f$(\textbf{ours}) & S & 2.15&0.04&75.59&60.88&57.91 \\
&
& 3PT$_{s00}f$(\textbf{ours}) & S & 1.99&0.04&76.94&62.66&\textbf{24.72} \\
&
& 4PT$_{suv}f$(M)~\cite{yu2025relative} & H~\cite{yu2025relative} & \underline{1.50}&\underline{0.03}&\underline{79.23}&\underline{66.34}&956.33 \\
&
& 3PT$_{s00}f$(\textbf{ours}) & H~\cite{yu2025relative} & \textbf{1.41}&\textbf{0.03}&\textbf{80.24}&\textbf{67.42}&967.52 \\
\cmidrule(rl){2- 9 }
& \multirow{ 6 }{*}{ \makecell{UniDepth \\ \cite{piccinelli2024unidepth}} }
& 3p3d~\cite{dingfundamental} & S & 3.49&0.07&69.47&55.57&\underline{27.57} \\
&
& 4PT$_{suv}f$(M)~\cite{yu2025relative} & S & 2.21&0.04&75.61&61.37&106.73 \\
&
& 4PT$_{suv}f$(\textbf{ours}) & S & 1.92&0.04&76.18&62.00&46.97 \\
&
& 3PT$_{s00}f$(\textbf{ours}) & S & 2.04&0.04&76.89&62.42&\textbf{23.47} \\
&
& 4PT$_{suv}f$(M)~\cite{yu2025relative} & H~\cite{yu2025relative} & \textbf{1.27}&\textbf{0.03}&\underline{81.68}&\textbf{69.64}&1107.04 \\
&
& 3PT$_{s00}f$(\textbf{ours}) & H~\cite{yu2025relative} & \underline{1.28}&\underline{0.03}&\textbf{81.99}&\underline{69.28}&1149.25 \\
\midrule

\multirow{ 31 }{*}{\makecell{ RoMA \\ \cite{edstedt2024roma} }}
& \multirow{ 1 }{*}{ \makecell{-} }
& 6PT~\cite{larsson2017efficient} & S & 1.15&0.02&85.23&75.03&147.48 \\
\cmidrule(rl){2- 9 }
& \multirow{ 6 }{*}{ \makecell{Real \\ Depth} }
& 3p3d~\cite{dingfundamental} & S & \textbf{0.93}&0.02&85.98&74.98&113.30 \\
&
& 4PT$_{suv}f$(M)~\cite{yu2025relative} & S & 0.99&\underline{0.02}&\underline{86.37}&\textbf{75.21}&167.49 \\
&
& 4PT$_{suv}f$(\textbf{ours}) & S & 1.03&0.02&86.25&\underline{75.04}&\underline{107.05} \\
&
& 3PT$_{s00}f$(\textbf{ours}) & S & \underline{0.99}&0.02&\textbf{86.68}&74.99&\textbf{79.96} \\
&
& 4PT$_{suv}f$(M)~\cite{yu2025relative} & H~\cite{yu2025relative} & 2.12&0.02&76.69&72.51&3218.22 \\
&
& 3PT$_{s00}f$(\textbf{ours}) & H~\cite{yu2025relative} & 2.28&\textbf{0.02}&76.42&72.43&3067.57 \\
\cmidrule(rl){2- 9 }
& \multirow{ 6 }{*}{ \makecell{MiDas \\ \cite{birkl2023midas}} }
& 3p3d~\cite{dingfundamental} & S & 1.91&0.02&78.71&68.10&86.60 \\
&
& 4PT$_{suv}f$(M)~\cite{yu2025relative} & S & \textbf{1.14}&\textbf{0.02}&\textbf{85.11}&\textbf{74.93}&142.78 \\
&
& 4PT$_{suv}f$(\textbf{ours}) & S & \underline{1.19}&\underline{0.02}&\underline{84.86}&\underline{74.80}&\underline{84.62} \\
&
& 3PT$_{s00}f$(\textbf{ours}) & S & 1.25&0.02&84.66&74.47&\textbf{65.79} \\
&
& 4PT$_{suv}f$(M)~\cite{yu2025relative} & H~\cite{yu2025relative} & 1.48&0.03&81.10&67.25&2301.37 \\
&
& 3PT$_{s00}f$(\textbf{ours}) & H~\cite{yu2025relative} & 1.48&0.03&81.44&67.03&2410.90 \\
\cmidrule(rl){2- 9 }
& \multirow{ 6 }{*}{ \makecell{DA v2 \\ \cite{yang2024depthv2}} }
& 3p3d~\cite{dingfundamental} & S & 2.07&0.02&78.86&68.29&91.39 \\
&
& 4PT$_{suv}f$(M)~\cite{yu2025relative} & S & \underline{1.21}&\underline{0.02}&\underline{85.33}&\underline{75.09}&146.35 \\
&
& 4PT$_{suv}f$(\textbf{ours}) & S & \textbf{1.20}&\textbf{0.02}&\textbf{85.61}&\textbf{75.43}&\underline{88.03} \\
&
& 3PT$_{s00}f$(\textbf{ours}) & S & 1.23&0.02&84.89&75.00&\textbf{67.61} \\
&
& 4PT$_{suv}f$(M)~\cite{yu2025relative} & H~\cite{yu2025relative} & 1.28&0.03&81.56&68.72&2161.36 \\
&
& 3PT$_{s00}f$(\textbf{ours}) & H~\cite{yu2025relative} & 1.30&0.03&81.57&68.68&2245.95 \\
\cmidrule(rl){2- 9 }
& \multirow{ 6 }{*}{ \makecell{MoGe \\ \cite{wang2024moge}} }
& 3p3d~\cite{dingfundamental} & S & 1.56&0.02&80.98&70.19&98.44 \\
&
& 4PT$_{suv}f$(M)~\cite{yu2025relative} & S & 1.02&0.02&85.91&75.61&151.65 \\
&
& 4PT$_{suv}f$(\textbf{ours}) & S & 1.10&0.02&85.85&75.80&\underline{92.92} \\
&
& 3PT$_{s00}f$(\textbf{ours}) & S & 1.05&0.02&86.04&75.83&\textbf{75.12} \\
&
& 4PT$_{suv}f$(M)~\cite{yu2025relative} & H~\cite{yu2025relative} & \textbf{0.89}&\underline{0.02}&\underline{86.99}&\textbf{76.66}&1923.31 \\
&
& 3PT$_{s00}f$(\textbf{ours}) & H~\cite{yu2025relative} & \underline{0.91}&\textbf{0.02}&\textbf{87.20}&\underline{76.50}&2043.91 \\
\cmidrule(rl){2- 9 }
& \multirow{ 6 }{*}{ \makecell{UniDepth \\ \cite{piccinelli2024unidepth}} }
& 3p3d~\cite{dingfundamental} & S & 1.89&0.02&82.50&71.89&97.99 \\
&
& 4PT$_{suv}f$(M)~\cite{yu2025relative} & S & 1.12&0.02&85.63&75.66&151.31 \\
&
& 4PT$_{suv}f$(\textbf{ours}) & S & 1.12&0.02&85.59&75.64&\underline{91.92} \\
&
& 3PT$_{s00}f$(\textbf{ours}) & S & 1.04&0.02&85.78&75.69&\textbf{75.62} \\
&
& 4PT$_{suv}f$(M)~\cite{yu2025relative} & H~\cite{yu2025relative} & \underline{0.83}&\textbf{0.02}&\underline{87.30}&\underline{77.34}&1997.88 \\
&
& 3PT$_{s00}f$(\textbf{ours}) & H~\cite{yu2025relative} & \textbf{0.82}&\underline{0.02}&\textbf{87.53}&\textbf{77.79}&2133.46 \\
\midrule
\multirow{ 8 }{*}{\makecell{ Mast3r \\ \cite{leroy2024grounding}} }
& \multirow{ 1 }{*}{ \makecell{-} }
& 6PT~\cite{larsson2017efficient} & S & 1.23&0.03&82.99&69.00&85.89 \\
\cmidrule(rl){2- 9 }
& \multirow{ 7 }{*}{\makecell{Mast3r \\ \cite{leroy2024grounding}}}
& 3p3d~\cite{dingfundamental} & S & 1.37&0.03&81.30&67.31&\underline{38.96} \\
&
& 4PT$_{suv}f$(M)~\cite{yu2025relative} & S & 1.58&0.03&79.68&66.05&95.89 \\
&
& 4PT$_{suv}f$(\textbf{ours}) & S & 1.36&0.03&80.86&67.16&48.36 \\
&
& 3PT$_{s00}f$(\textbf{ours}) & S & \underline{1.34}&\underline{0.03}&\underline{82.09}&\underline{68.26}&\textbf{34.54} \\
&
& 4PT$_{suv}f$(M)~\cite{yu2025relative} & H~\cite{yu2025relative} & 2.43&0.05&72.28&58.32&3825.11 \\
&
& 3PT$_{s00}f$(\textbf{ours}) & H~\cite{yu2025relative} & 2.39&0.04&72.45&58.61&4079.21 \\
&
& - & M~\cite{leroy2024grounding} & \textbf{1.32}&\textbf{0.01}&\textbf{85.64}&\textbf{82.95}&4800.37 \\
\hline
\end{tabular}}
    \caption{Comparison of different methods on the \ETH dataset~\cite{schops2017multi} for the equal and unknown focal length case. Opt.: S - PoseLib~\cite{poselib} implementation using Sampson error, H - hybrid RANSAC from~\cite{yu2025relative}, M - non-linear optimization used in~\cite{leroy2024grounding}.}
    \label{tab:real_equal}
    \vspace{-5mm}
\end{table}

\subsection{Different Focal Lengths}
\begin{figure}
    \setlength{\tabcolsep}{0pt} 
    \centering
    % \begin{tabular}
    % {m{0.68\linewidth}m{0.28\linewidth}}
    \resizebox{\linewidth}{!}{    
    \begin{tikzpicture} 
        \begin{axis}[%
        hide axis, xmin=0,xmax=0,ymin=0,ymax=0,
        legend style={draw=white!15!white, 
        line width = 1pt,
        legend cell align=left,
        legend  columns =5, % comment for column display
        /tikz/every even column/.append style={column sep=0.05cm},
        font=\scriptsize
        },
        legend image post style={xscale=1}
        ]
        \addlegendimage{Seaborn1}        \addlegendentry{3PT$_{s00}f_{1,2}$(\textbf{ours})};
        \addlegendimage{Seaborn2}        \addlegendentry{4PT$_{suv}f_{1,2}$(\textbf{ours})};        
        \addlegendimage{Seaborn3}        \addlegendentry{4PT$_{suv}f_{1,2}$(M)~\cite{yu2025relative}};

        \addlegendimage{Seaborn4}        \addlegendentry{4p4d~\cite{dingfundamental}};        
        \addlegendimage{Seaborn6}
        \addlegendentry{7PT};
        \end{axis}
    \end{tikzpicture}}
    \includegraphics[width=0.8\linewidth]{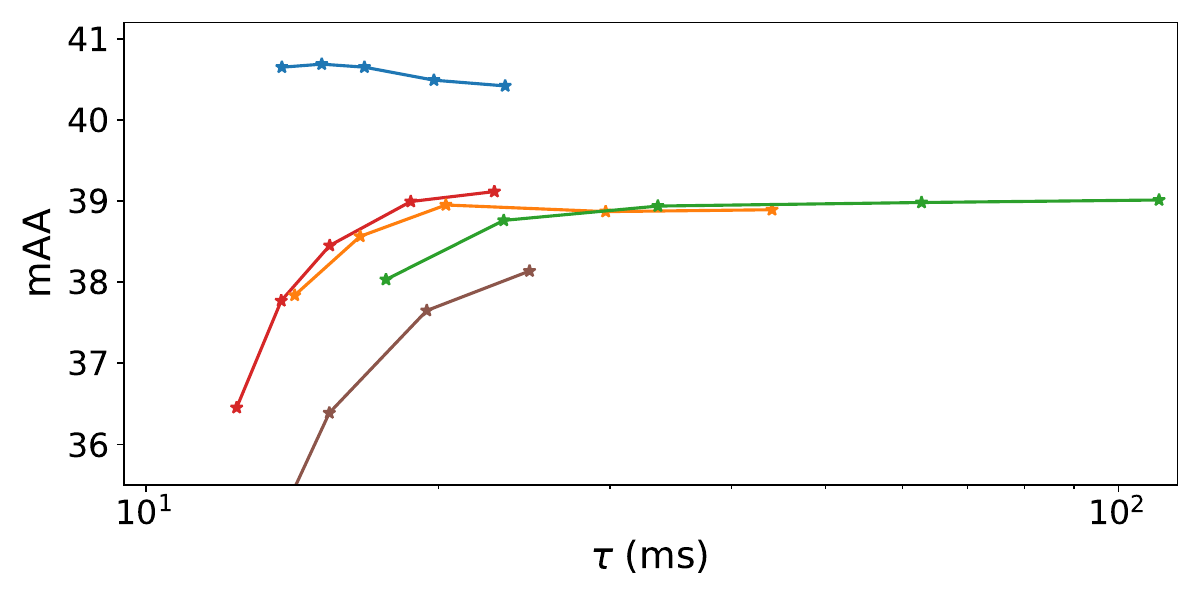}        
    % \end{tabular}
    \vspace{-3mm}
    \caption{Speed accuracy evaluation for different unknown focal length case using the SP+LG matches~\cite{detone2018superpoint, lindenberger2023lightglue} and MoGe~\cite{wang2024moge} depths on the \Phototourism dataset~\cite{Jin2020}. We evaluated mAA and runtimes ($\tau$) for each method running for 50, 100, 200, 500 and 1000 iterations within PoseLib~\cite{poselib}.}
    \label{fig:varying_speed_acc_splg_pt}
    \vspace{-6mm}
\end{figure}

Table~\ref{tab:real_varying} shows the comparison of the methods on the \Phototourism dataset for two cameras with different unknown focal lengths. As for the previous two cases, when good depth estimate is available the best results are obtained with the hybrid RANSAC scheme~\cite{yu2025relative} in which our solver 3PT$_{s00}f_{1,2}$ outperforms 4PT$_{suv}f_{1,2}$(M)~\cite{yu2025relative} showing that it is not necessary to model shift for good performance. This is also confirmed by the experiments in PoseLib~\cite{poselib} which show that 3PT$_{s00}f$ outperforms other depth-based solvers across all combinations of matches and MDEs in terms of pose accuracy while also being significantly faster than solvers which incorporate shift. 
%Additionally, 
Figure~\ref{fig:varying_speed_acc_splg_pt} shows that 3PT$_{s00}f$ clearly outperforms the alternatives in terms of speed and pose estimation accuracy. We observe a slight performance drop in our 3PT$_{s00}f$ solver as the number of iterations increases. This may be attributed to the solver converging on physically impossible solutions that yield better scores, such as unrealistic scales or focal lengths. Our current implementation does not include checks to validate the physical feasibility of these parameters (we only consider positive scale and focal length). Incorporating filtering of impossible solutions could be a future work.

\begin{table}[]
    \centering
      \resizebox{\linewidth}{!}{
\begin{tabular}{cclcccccc}
\toprule
% \multirow{2.5}{*}{{Matches}} & \multirow{2.5}{*}{{Depth}} &  \multirow{2.5}{*}{{Solver}} & \multirow{2.5}{*}{{Opt}} & \multicolumn{5}{c}{Phototourism}\\
% \cmidrule(rl){5-9}
% & & & & $\epsilon(^\circ)\downarrow$ & $\epsilon_f(^\circ)\downarrow$ & mAA$\uparrow$ & mAA$_f\uparrow$ & $\tau (ms)\downarrow$ \\
\multirow{1}{*}{{Matches}} & \multirow{1}{*}{{Depth}} &  \multirow{1}{*}{{Solver}} & \multirow{1}{*}{{Opt}} & $\epsilon(^\circ)\downarrow$ & $\epsilon_f(^\circ)\downarrow$ & mAA$\uparrow$ & mAA$_f\uparrow$ & $\tau (ms)\downarrow$ \\
\midrule
\multirow{ 31 }{*}{ \makecell{SP+LG \\ \cite{detone2018superpoint, lindenberger2023lightglue}} }
& \multirow{ 1 }{*}{ \makecell{-} }
& 7PT~\cite{hartley2003multiple} & S & 8.03&0.17&38.14&23.78&24.80 \\
\cmidrule(rl){2- 9 }
& \multirow{ 6 }{*}{ \makecell{Real \\ Depth} }
& 4p4d~\cite{dingfundamental} & S & 7.58&0.16&39.34&24.65&\textbf{23.69} \\
&
& 4PT$_{suv}f_{1,2}$(M)~\cite{yu2025relative} & S & 7.33&0.17&39.56&24.08&118.98 \\
&
& 4PT$_{suv}f_{1,2}$(\textbf{ours}) & S & 7.40&0.17&39.55&24.13&36.71 \\
&
& 3PT$_{s00}f_{1,2}$(\textbf{ours}) & S & 6.74&0.15&41.25&24.60&\underline{25.27} \\
&
& 4PT$_{suv}f_{1,2}$(M)~\cite{yu2025relative} & H~\cite{yu2025relative} & \underline{2.86}&\underline{0.04}&\underline{63.88}&\underline{53.46}&1922.12 \\
&
& 3PT$_{s00}f_{1,2}$(\textbf{ours}) & H~\cite{yu2025relative} & \textbf{2.82}&\textbf{0.04}&\textbf{64.19}&\textbf{53.73}&1901.23 \\
\cmidrule(rl){2- 9 }
& \multirow{ 6 }{*}{ \makecell{MiDas \\ \cite{birkl2023midas}} }
& 4p4d~\cite{dingfundamental} & S & 12.50&0.24&30.95&18.92&\underline{20.67} \\
&
& 4PT$_{suv}f_{1,2}$(M)~\cite{yu2025relative} & S & \underline{8.75}&\underline{0.19}&\underline{36.60}&\textbf{22.37}&112.19 \\
&
& 4PT$_{suv}f_{1,2}$(\textbf{ours}) & S & 8.91&0.19&36.25&\underline{22.18}&32.32 \\
&
& 3PT$_{s00}f_{1,2}$(\textbf{ours}) & S & \textbf{8.60}&\textbf{0.19}&\textbf{36.62}&21.94&\textbf{18.17} \\
&
& 4PT$_{suv}f_{1,2}$(M)~\cite{yu2025relative} & H~\cite{yu2025relative} & 13.29&0.27&21.68&13.55&2599.72 \\
&
& 3PT$_{s00}f_{1,2}$(\textbf{ours}) & H~\cite{yu2025relative} & 13.40&0.28&21.33&13.39&2574.24 \\
\cmidrule(rl){2- 9 }
& \multirow{ 6 }{*}{ \makecell{DA v2 \\ \cite{yang2024depthv2}} }
& 4p4d~\cite{dingfundamental} & S & 10.40&0.20&34.72&22.08&\underline{22.02} \\
&
& 4PT$_{suv}f_{1,2}$(M)~\cite{yu2025relative} & S & \underline{8.17}&\underline{0.18}&37.62&\underline{23.32}&113.20 \\
&
& 4PT$_{suv}f_{1,2}$(\textbf{ours}) & S & 8.19&0.18&\underline{37.62}&23.27&33.14 \\
&
& 3PT$_{s00}f_{1,2}$(\textbf{ours}) & S & \textbf{7.66}&\textbf{0.16}&\textbf{38.90}&\textbf{23.57}&\textbf{20.44} \\
&
& 4PT$_{suv}f_{1,2}$(M)~\cite{yu2025relative} & H~\cite{yu2025relative} & 9.55&0.21&32.88&20.88&2490.39 \\
&
& 3PT$_{s00}f_{1,2}$(\textbf{ours}) & H~\cite{yu2025relative} & 9.48&0.21&32.49&20.65&2494.66 \\
\cmidrule(rl){2- 9 }
& \multirow{ 6 }{*}{ \makecell{MoGe \\ \cite{wang2024moge}} }
& 4p4d~\cite{dingfundamental} & S & 7.71&0.16&39.12&24.44&\textbf{22.82} \\
&
& 4PT$_{suv}f_{1,2}$(M)~\cite{yu2025relative} & S & 7.54&0.17&39.01&23.83&110.08 \\
&
& 4PT$_{suv}f_{1,2}$(\textbf{ours}) & S & 7.65&0.17&38.89&23.85&44.03 \\
&
& 3PT$_{s00}f_{1,2}$(\textbf{ours}) & S & \underline{7.03}&0.16&40.42&24.43&\underline{23.41} \\
&
& 4PT$_{suv}f_{1,2}$(M)~\cite{yu2025relative} & H~\cite{yu2025relative} & 7.13&\underline{0.11}&\underline{46.38}&\underline{34.35}&857.73 \\
&
& 3PT$_{s00}f_{1,2}$(\textbf{ours}) & H~\cite{yu2025relative} & \textbf{5.90}&\textbf{0.08}&\textbf{49.14}&\textbf{37.80}&840.92 \\
\cmidrule(rl){2- 9 }
& \multirow{ 6 }{*}{ \makecell{UniDepth \\ \cite{piccinelli2024unidepth}} }
& 4p4d~\cite{dingfundamental} & S & 7.67&0.16&39.18&24.61&\textbf{23.63} \\
&
& 4PT$_{suv}f_{1,2}$(M)~\cite{yu2025relative} & S & 7.53&0.17&39.25&23.98&118.02 \\
&
& 4PT$_{suv}f_{1,2}$(\textbf{ours}) & S & 7.55&0.17&39.18&23.99&36.04 \\
&
& 3PT$_{s00}f_{1,2}$(\textbf{ours}) & S & 7.01&0.15&40.51&24.48&\underline{24.61} \\
&
& 4PT$_{suv}f_{1,2}$(M)~\cite{yu2025relative} & H~\cite{yu2025relative} & \underline{3.38}&\underline{0.06}&\underline{60.15}&\underline{47.02}&1945.50 \\
&
& 3PT$_{s00}f_{1,2}$(\textbf{ours}) & H~\cite{yu2025relative} & \textbf{3.35}&\textbf{0.06}&\textbf{60.38}&\textbf{47.24}&1957.65 \\
\midrule
\multirow{ 31 }{*}{ \makecell{RoMA \\ \cite{edstedt2024roma}}}
& \multirow{ 1 }{*}{ \makecell{-} }
& 7PT~\cite{hartley2003multiple} & S & 4.30&0.10&53.16&34.73&75.10 \\
\cmidrule(rl){2- 9 }
& \multirow{ 6 }{*}{ \makecell{Real \\ Depth} }
& 4p4d~\cite{dingfundamental} & S & 4.24&0.10&53.51&35.01&\textbf{77.49} \\
&
& 4PT$_{suv}f_{1,2}$(M)~\cite{yu2025relative} & S & 4.22&0.10&54.07&34.83&190.34 \\
&
& 4PT$_{suv}f_{1,2}$(\textbf{ours}) & S & 4.24&0.10&53.97&34.84&107.46 \\
&
& 3PT$_{s00}f_{1,2}$(\textbf{ours}) & S & 4.08&0.10&55.22&35.05&\underline{100.87} \\
&
& 4PT$_{suv}f_{1,2}$(M)~\cite{yu2025relative} & H~\cite{yu2025relative} & \underline{1.60}&\underline{0.03}&\underline{75.10}&\underline{64.07}&3518.89 \\
&
& 3PT$_{s00}f_{1,2}$(\textbf{ours}) & H~\cite{yu2025relative} & \textbf{1.58}&\textbf{0.03}&\textbf{75.41}&\textbf{64.37}&3531.28 \\
\cmidrule(rl){2- 9 }
& \multirow{ 6 }{*}{ \makecell{MiDas \\ \cite{birkl2023midas}} }
& 4p4d~\cite{dingfundamental} & S & 5.58&0.13&46.41&29.53&\textbf{67.95} \\
&
& 4PT$_{suv}f_{1,2}$(M)~\cite{yu2025relative} & S & 4.96&0.12&\underline{51.13}&\textbf{33.29}&164.23 \\
&
& 4PT$_{suv}f_{1,2}$(\textbf{ours}) & S & \underline{4.91}&\underline{0.12}&51.09&\underline{33.08}&84.29 \\
&
& 3PT$_{s00}f_{1,2}$(\textbf{ours}) & S & \textbf{4.70}&\textbf{0.11}&\textbf{51.39}&32.56&\underline{71.37} \\
&
& 4PT$_{suv}f_{1,2}$(M)~\cite{yu2025relative} & H~\cite{yu2025relative} & 9.10&0.21&32.05&19.20&4662.34 \\
&
& 3PT$_{s00}f_{1,2}$(\textbf{ours}) & H~\cite{yu2025relative} & 9.11&0.21&32.04&19.30&4680.00 \\
\cmidrule(rl){2- 9 }
& \multirow{ 6 }{*}{ \makecell{DA v2 \\ \cite{yang2024depthv2}} }
& 4p4d~\cite{dingfundamental} & S & 4.61&0.11&51.24&33.30&\textbf{73.93} \\
&
& 4PT$_{suv}f_{1,2}$(M)~\cite{yu2025relative} & S & 4.43&\underline{0.10}&\underline{52.90}&\underline{34.36}&170.68 \\
&
& 4PT$_{suv}f_{1,2}$(\textbf{ours}) & S & \underline{4.42}&0.10&52.83&34.35&90.36 \\
&
& 3PT$_{s00}f_{1,2}$(\textbf{ours}) & S & \textbf{4.20}&\textbf{0.10}&\textbf{53.95}&\textbf{34.56}&\underline{81.59} \\
&
& 4PT$_{suv}f_{1,2}$(M)~\cite{yu2025relative} & H~\cite{yu2025relative} & 6.53&0.16&43.25&27.71&4026.79 \\
&
& 3PT$_{s00}f_{1,2}$(\textbf{ours}) & H~\cite{yu2025relative} & 6.52&0.16&43.30&27.79&4113.61 \\
\cmidrule(rl){2- 9 }
& \multirow{ 6 }{*}{ \makecell{MoGe \\ \cite{wang2024moge}} }
& 4p4d~\cite{dingfundamental} & S & 4.22&0.10&53.61&34.99&\textbf{77.14} \\
&
& 4PT$_{suv}f_{1,2}$(M)~\cite{yu2025relative} & S & 4.33&0.10&53.44&34.72&180.83 \\
&
& 4PT$_{suv}f_{1,2}$(\textbf{ours}) & S & 4.34&0.10&53.42&34.71&100.14 \\
&
& 3PT$_{s00}f_{1,2}$(\textbf{ours}) & S & 4.18&0.10&54.49&34.98&\underline{95.72} \\
&
& 4PT$_{suv}f_{1,2}$(M)~\cite{yu2025relative} & H~\cite{yu2025relative} & \underline{2.56}&\underline{0.05}&\underline{68.05}&\underline{51.63}&2096.95 \\
&
& 3PT$_{s00}f_{1,2}$(\textbf{ours}) & H~\cite{yu2025relative} & \textbf{2.50}&\textbf{0.05}&\textbf{68.57}&\textbf{52.03}&2158.75 \\
\cmidrule(rl){2- 9 }
& \multirow{ 6 }{*}{ \makecell{UniDepth \\ \cite{piccinelli2024unidepth}} }
& 4p4d~\cite{dingfundamental} & S & 4.20&0.10&53.84&35.24&\textbf{77.78} \\
&
& 4PT$_{suv}f_{1,2}$(M)~\cite{yu2025relative} & S & 4.30&0.10&53.60&34.85&182.14 \\
&
& 4PT$_{suv}f_{1,2}$(\textbf{ours}) & S & 4.29&0.10&53.53&34.71&99.76 \\
&
& 3PT$_{s00}f_{1,2}$(\textbf{ours}) & S & 4.18&0.10&54.39&34.91&\underline{96.84} \\
&
& 4PT$_{suv}f_{1,2}$(M)~\cite{yu2025relative} & H~\cite{yu2025relative} & \underline{2.30}&\underline{0.04}&\underline{69.95}&\underline{53.90}&3440.95 \\
&
& 3PT$_{s00}f_{1,2}$(\textbf{ours}) & H~\cite{yu2025relative} & \textbf{2.29}&\textbf{0.04}&\textbf{70.18}&\textbf{54.09}&3533.26 \\
\midrule
\multirow{ 8 }{*}{ \makecell{Mast3r \\ \cite{leroy2024grounding} }}
& \multirow{ 1 }{*}{ \makecell{-} }
& 7PT~\cite{hartley2003multiple} & S & 4.39&0.09&54.01&35.02&39.53 \\
\cmidrule(rl){2- 9 }
& \multirow{ 7 }{*}{\makecell{Mast3r \\ \cite{leroy2024grounding}}}
& 4p4d~\cite{dingfundamental} & S & \underline{5.20}&\underline{0.11}&\underline{49.69}&\underline{31.64}&\textbf{36.90} \\
&
& 4PT$_{suv}f_{1,2}$(M)~\cite{yu2025relative} & S & 7.02&0.14&43.95&27.62&108.64 \\
&
& 4PT$_{suv}f_{1,2}$(\textbf{ours}) & S & 7.03&0.14&44.10&27.81&44.88 \\
&
& 3PT$_{s00}f_{1,2}$(\textbf{ours}) & S & 5.40&0.12&49.25&30.88&\underline{37.40} \\
&
& 4PT$_{suv}f_{1,2}$(M)~\cite{yu2025relative} & H~\cite{yu2025relative} & 10.11&0.24&35.16&20.56&4871.24 \\
&
& 3PT$_{s00}f_{1,2}$(\textbf{ours}) & H~\cite{yu2025relative} & 10.13&0.25&35.17&20.51&4980.42 \\
&
& - & M~\cite{leroy2024grounding} & \textbf{2.71}&\textbf{0.04}&\textbf{66.54}&\textbf{56.43}&4903.10 \\
\hline
\end{tabular}}
    \caption{Comparison of different methods on the \Phototourism dataset~\cite{Jin2020} for the two unknown focal length case. Opt.: S - PoseLib~\cite{poselib} implementation using Sampson error, H - hybrid RANSAC from~\cite{yu2025relative}, M - non-linear optimization used in~\cite{leroy2024grounding}.}
    \label{tab:real_varying}
    \vspace{-2mm}
\end{table}

\noindent\textbf{Additional Experiments} We provide additional experiments in SM which in addition to results on additional datasets also provide evaluation of alternative scoring and LO strategies within PoseLib as well as the proposed 3PT$_{100}f$ and 3PT$_{100}f_{1,2}$ solvers. We also show that it is possible to use Mast3r~\cite{leroy2024grounding} for matches in combination with MDE to improve estimated poses and intrinsics.

\noindent\textbf{Limitations.}  
%
%Depth-based solvers need additional computation time for generating depth. However, note that in many applications, mono-depth estimation methods are also used for other tasks, and thus this computational overhead is not purely introduced by the pose estimation method. In addition, 
Although using depth maps does not add computational overhead in pose estimation, obtaining these depth maps using learning-based methods introduces additional costs. This is usually a few milliseconds per image~\cite{yu2025relative}. However, in many applications, MDE methods are used for multiple tasks. Thus, the computational overhead is not solely attributable to the pose estimation method.

Compared to standard point-based focal length solvers, the proposed solvers encounter fewer degenerate cases. The formulation allows us to handle pure rotation effectively. However, there are still several degenerate cases similar to those in point-based solvers; for example, four coplanar points and cases of pure translation can introduce degeneracies in focal length recovery~\cite{kahl1999critical}.
\section{Conclusion}

We address the problem of estimating the relative pose of two cameras using monocular depth predictions.
%for both images. 
Unlike prior work, which only considered relative depths and their unknown scale factors, we modeled the fact that depth maps can be defined up to unknown scale and shift parameters. 
We propose multiple solvers that jointly estimate the scale, shift (or only one of them), and the relative pose. 
We consider
%all variants 
solvers for calibrated cameras, cameras with shared unknown or different unknown focal lengths. 
Efficient solvers that outperform state-of-the-art depth-aware solvers are proposed for all three cases. In extensive experiments, we discuss which solvers are preferable in different situations \eg for precise or imprecise depths. 
%and experiments on real and synthetic data show the practical relevance of our solvers. 
%In particular, we evaluate our novel solvers on three real-world datasets and 5 depth prediction networks, showing that our solvers achieve state-of-the-art results.

% \noindent\textbf{Acknowledgements}
% Y. D., V.V. and Z.K. were supported by the Czech Science Foundation (GAČR) JUNIOR STAR Grant (No. 22-23183M). V. K. was supported by the project no. 1/0373/23. and the TERAIS project, a Horizon-Widera-2021 program of the European Union under the Grant agreement number 101079338. T. S. was supported by the EU Horizon 2020 project RICAIP (grant agreement No. 857306).

{
\small
\bibliographystyle{ieeenat_fullname}
\bibliography{main}
}

 \clearpage
 \setcounter{page}{1}
 \maketitlesupplementary

% \appendix

% \noindent\textbf{Appendices}

% \begin{abstract}
This supplementary material provides the following information: 
	Sec.~\ref{sec:supp_solvers} provides more details about the proposed solvers, including a general approach that can be used to solve all variants of the depth-aware relative pose problem that require 3 point correspondences, the  3PT$_{suv}$(inverse) solver for affine-invariant inverse depths, and the variants of the affine-invariant 4PT focal length solvers. Sec.~\ref{sec:supp_results} provides more experimental results.

\section{More Details About the Solvers}\label{sec:supp_solvers}

\subsection{Solvers Using Three Point Correspondences}
For calibrated camera pose estimation with monocular depth, all possible cases can be solved using three point correspondences and a varying number of monocular depth estimates. Similarly, for focal length problems, most cases can be solved using three point correspondences and a varying number of depth estimates. % different monocular depth values. 
In general, all cases that involve three point correspondences can be solved using a similar approach.

Here we show the solution to the shared unknown focal length scale-invariant case, \ie the 3PT$_{s00}f$ solver.
In this case, the shifts in the monocular depths are omitted (considered to be zero) and we only consider the unknown scales. The minimal case is two 3D-3D point correspondence with one 3D-2D point correspondences. 
We have
\begin{equation}
\begin{split}
    \|s\M K^{-1}(\beta_1{\M q}_1 - \beta_2{\M q}_2) \|  &=\|\M K^{-1} (\alpha_1{\M p}_1 - \alpha_2{\M p}_2) \|, \nonumber \\
\|s\M K^{-1}(\beta_1{\M q}_1 - \eta_3{\M q}_3) \|  &=\|\M K^{-1} (\alpha_1{\M p}_1 - \alpha_3{\M p}_3) \|,\label{eq:s01} \\
\|s\M K^{-1}(\beta_2{\M q}_2 - \eta_3{\M q}_3) \|  &=\|\M K^{-1} (\alpha_2{\M p}_2 - \alpha_3{\M p}_3) \|.  \nonumber
\end{split}
\end{equation}
where $\alpha_1,\alpha_2,\alpha_3,\beta_1,\beta_2$ are known depths estimated \eg using MDE network, and $\eta_3$ is the unknown depth.\footnote{
Note that in this case, for the last (third) correspondence, we assume that we know/use the depth only from one image, \ie we have 3D-2D correspondence with unknown depth $\eta_3$.}
There are three equations in three unknowns $\{s,f,\eta_3\}$, which can be solved similarly as for the 3PT$_{suv}$ solver presented in Sec.~3.1 of the main paper. In general, all the problems using three point correspondences can be converted into solving three equations in three unknowns. 
They differ in %The difference is using different 
the number of depth parameters for the 2D points, but the structure and the solution strategy in all cases similar.

\subsection{ 3PT\texorpdfstring{$_{suv}$}{suv} Inverse Depth Solver}

\noindent
Some MDE networks return affine-invariant inverse depths. In this case, the true depths can be expressed as
\begin{equation}
% \begin{split}
\eta_i = \frac{s_1}{\alpha_{i} + u}  ,\ \lambda_i = \frac{s_2}{\beta_{i} + v} , \label{eq:s02}
% \end{split}
\end{equation}
where $\alpha_{i},\beta_{i}$ are known values from the inverse monocular depth, and $\{s_1,s_2\},\{u,v\}$ are the unknown scales and shifts in the inverse depth. In this case, we have
% \begin{equation}
% \frac{s_2}{\beta_{i} + v}\tilde{\M q}_i  =  \frac{s_1}{\alpha_{i} + u} \M R \tilde{\M p}_i  +\M T, \label{eq:06}
% \end{equation}
\begin{equation}
\frac{s_2}{\beta_{i} + v}\M K_2^{-1} \M q_{i}  =  \frac{s_1}{\alpha_{i} + u} \M R \M K_1^{-1} \M p_{i}  +\M T, \label{eq:s03}
\end{equation}
Dividing~\eqref{eq:s03} by $s_1$ gives
\begin{equation}
\frac{s}{\beta_{i} + v}\M K_2^{-1} \M q_{i} =  \frac{1}{\alpha_{i} + u} \M R \M K_1^{-1} \M p_{i} +\M t, \label{eq:s04}.
\end{equation}
In this case, similarly to the affine-invariant depth case, we have 9 DOF for calibrated cameras.
However, in contrast to the affine-invariant depths, the constraints~\eqref{eq:s04} for affine-invariant inverse depths are more complicated, since they contain unknown parameters in the denominators. 
We can use similar tricks to eliminate the rotation and translation from the original equations~\eqref{eq:s04} as the ones used for the affine-invariant depth solvers presented in the main paper. In this case, we obtain
\begin{equation}
\begin{aligned}
\|\frac{s\tilde{\M q}_1}{\beta_1+v} - \frac{s\tilde{\M q}_2}{\beta_2+v} \|  =\| \frac{\tilde{\M p}_1}{\alpha_1+u} - \frac{\tilde{\M p}_2}{\alpha_2+u}\M  \|, \\
\|\frac{s\tilde{\M q}_1}{\beta_1+v} - \frac{s\tilde{\M q}_3}{\beta_3+v} \|  =\| \frac{\tilde{\M p}_1}{\alpha_1+u} - \frac{\tilde{\M p}_3}{\alpha_3+u}\M  \|, \\
\|\frac{s\tilde{\M q}_2}{\beta_2+v} - \frac{s\tilde{\M q}_3}{\beta_3+v} \|  =\| \frac{\tilde{\M p}_2}{\alpha_2+u} - \frac{\tilde{\M p}_3}{\alpha_3+u}\M  \|.
\end{aligned}
\label{eq:s05}
\end{equation}
However, these equations have unknowns in the denominators, and simply multiplying the equations with the denominators results in a very complex system of equations that is difficult to solve. 

To solve the equations efficiently, we first multiply~\eqref{eq:s05} with $\alpha_1 + u$, and let 
% \begin{equation}
\begin{align}
    b_1 &= \frac{s(\alpha_1 + u)}{\beta_1 + v},\ b_2 = \frac{s(\alpha_1 + u)}{\beta_2 + v},\ b_3 = \frac{s(\alpha_1 + u)}{\beta_3 + v}, \nonumber \\
    \ c_2 &= \frac{\alpha_1 + u}{\alpha_2 + u},\ c_3 = \frac{\alpha_1 + u}{\alpha_3 + u}.\label{eq:s06}
\end{align}
% \end{equation}
Substituting~\eqref{eq:s06} into~\eqref{eq:s05} we have three equations
\begin{equation}
\begin{aligned}
\|b_1\tilde{\M q}_1 - b_2\tilde{\M q}_2 \|  &=\| \tilde{\M p}_1 - c_2\tilde{\M p}_2 \|, \\
\|b_1\tilde{\M q}_1 - b_3\tilde{\M q}_3 \|  &=\|\tilde{\M p}_1 - c_3\tilde{\M p}_3 \|, \\
\|b_2\tilde{\M q}_2 - b_3\tilde{\M q}_3 \|  &=\|c_2\tilde{\M p}_2 - c_3\tilde{\M p}_3 \|, 
\end{aligned}\label{eq:s07}
\end{equation}
where $b_1,b_2,b_3,c_2,c_3$ are new unknowns. However, these unknown are not independent.
To find the constraints on $b_1,b_2,b_3,c_2,c_3$, we use the elimination ideal technique~\cite{cox2006using}.
In this case, we first create an ideal $J$ generated by five polynomials~\eqref{eq:s06}. Then, the unknown parameters $s,u,v$ are eliminated from the generators of $J$ by computing the generators of the elimination ideal $J_1 = J \cap \mathbb{C}[\alpha_1, \alpha_2 , ..., c_2, c_3]$. These generators can be computed using the following Macaulay2~\cite{M2} code 
{
\small
\begin{lstlisting}
R = QQ[s,u,v,$\alpha_1$,$\alpha_2$,$\alpha_3$,$\beta_1$,$\beta_2$,$\beta_3$,$b_1$,$b_2$,$b_3$,$c_2$,$c_3$];
eq = {$b_1(\beta_1+v)-s(\alpha_1+u)$, $b_2(\beta_2+v)-s(\alpha_1+u)$,
      $b_3(\beta_3+v)-s(\alpha_1+u)$, $c_2(\alpha_2+u)-(\alpha_1+u)$,
      $c_3(\alpha_3+u)-(\alpha_1+u)$};
J = ideal(eq);
J1 = eliminate(J,{s,u,v});
g = mingens J1;
"constraints.txt" << toString g << close;
\end{lstlisting}
}% 
\noindent In this case, by eliminating $\{s,u,v\}$ from~\eqref{eq:s06} we obtain the following two equations in $\{b_1,b_2,b_3,c_2,c_3\}$
\begin{equation}
\resizebox{0.88\hsize}{!}{%
$
    \begin{aligned}
        &b_1b_2\beta_1-b_1b_3\beta_1-b_1b_2\beta_2+b_2b_3\beta_2+b_1b_3\beta_3-b_2b_3\beta_3=0, \\
&c_2c_3\alpha_2-c_2c_3\alpha_3+c_2\alpha_1-c_3\alpha_1-c_2\alpha_2+c_3\alpha_3=0.
    \end{aligned}$
    }\label{eq:s08}
\end{equation}
Combining~\eqref{eq:s08} with~\eqref{eq:s07} we have 5 equations in 5 unknowns, which can be solved using the Gr\"{o}bner basis method~\cite{cox2006using}. Using the automatic generator of Gr\"{o}bner basis solvers~\cite{larsson2017efficient}, we obtain a solver with an elimination template of size $54\times 66$ and 12 solutions. Note that there are two trivial solutions $b_2=b_3=c_2=c_3=0, \|b_1 \tilde{\M q}_1 \| = \|\tilde{\M p}_1 \|$.
%, which should be eliminated. 
Hence, there are up to 10 feasible solutions. 

The 3PT$_{suv}$(inverse) solver is much more complex than the 3PT$_{suv}$ solver for affine-invariant depths presented in the main paper. In the next section, we show that the 3PT$_{suv}$(inverse) solver does not give better results than the 3PT$_{suv}$ solver inside RANSAC even when used with affine-invariant inverse depths.

\subsection{Fast 4PT Solvers}

In Sec 3.2 of the main paper, we have mentioned that the focal length problems with affine-invariant depth can be efficiently solved using all the six equations. Here we provide more details on the solutions.

\noindent\textbf{4PT$_{suv}f$(Eigen).} By using four 3D-3D point correspondences, 
%and all the six equations, 
we can rewrite the six equations for this problem as
\begin{equation}
    \M M\ [1,c,cv,cv^2,u,u^2,f^2,cf^2]^\top =0,
    %\ i=1,2,..,6,
    \label{eq:s09}
\end{equation}
where $\M M$ is a $6\times 8$ coefficient matrix. 

Since these equations only contain $f^2$, we let $w=f^2$ and consider $w$ as the hidden variable~\cite{kukelova2012polynomial}. Then~\eqref{eq:s09} can be written as
\begin{equation}
    \M M(w)\ [1,c,cv,cv^2,u,u^2]^\top =0,
    \label{eq:s10}
\end{equation}
where $\M M(w)$ is a $6\times 6$ polynomial matrix in $w$. In this case
\begin{equation}
    \M M(w) = \M M_0 + w\M M_1,
    \label{eq:s11}
\end{equation}
where $\M M_0$ and  $\M M_1$  are $6 \times 6$ coefficient matrices.

Thus, in this case, the solutions to $1/w$ are the eigenvalues of the following matrix
\begin{equation}
\M A = 
 -{\M M}_0^{-\top}{\M M}_1^\top .\label{eq:s12}
\end{equation}
Note that there are 4 zero columns in $\M M_1$, which will result in zero eigenvalues. Based on~\cite{kukelova2012polynomial}, these zero columns can be removed together with the zero rows. Hence, we only need to find the eigenvalues of a $2\times 2$ matrix resulting in 2 solutions to the problem. We denote this solver as 4PT$_{suv}f$(Eigen).

\vspace{1mm}
\noindent\textbf{4PT$_{suv}f_{1,2}$(Eigen).} For different and unknown focal lengths case, we have the following six equations
\begin{equation}
    \M m_i\ [1,c,cf_2^2,cf_2^2v,cf_2^2v^2,f_1^2,f_1^2u,f_1^2u^2]^\top =0,
    \label{eq:s13}
\end{equation}
where $i=1,2,...,6$. We consider $v$ as a hidden variable, and~\eqref{eq:s13} can be written as
\begin{equation}
    \M M(v)\ [1,c,cf_2^2,f_1^2,f_1^2u,f_1^2u^2]^\top =0,
    \label{eq:s14}
\end{equation}
where $\M M(v)$ is a $6\times 6$ polynomial matrix in $v$. It can be solved similarly to the shared unknown focal length case, and there are only two possible solutions. We denote this solver as 4PT$_{suv}f_{1,2}$(Eigen).

\section{More Experiments}\label{sec:supp_results}

% \subsection{Stability Under Pure Rotation}
% Figure~\ref{fig:supp_stability} evaluates the numerical stability of the focal length solvers in the case of a pure rotation between the two cameras. 
% By construction, the focal length solvers can handle pure rotation since we first eliminate the translation, \ie, the constraints used to solve the problems are independent of translation. 
% This can also be seen from the results of the experiment.
% %This formulation, therefore, enables us to handle pure rotation. 
% Note that if we have prior knowledge of the motion, \ie, if we know that the cameras are undergoing pure rotation, then fewer point correspondences can be used to solve all the problems. 
% \begin{figure}[t]
% \begin{tabular}[b]{c} 
% \subfloat[]{\includegraphics[width=0.46\columnwidth]{figures_supp/stab_focal_p_rot.eps}}\; 
% \subfloat[]{\includegraphics[width=0.46\columnwidth]{figures_supp/stab_focal_f_rot.eps}}\\
% 	\end{tabular}
% 	\caption{Numerical stability of the focal length solvers under pure rotation. We report (a) the rotation error, and (b) the focal length error.
%  }
% \label{fig:supp_stability}
% \end{figure}

\subsection{Results for 3PT\texorpdfstring{$_{suv}$}{suv}(Inverse)}

This solver was derived to be used with affine-invariant inverse depths,~\eg, obtained via Depth Anything~\cite{yang2024depth}. However, we observed that the 3PT$_{suv}$(inverse) solver does not improve the accuracy even for affine-invariant inverse depths when used inside RANSAC. In addition, 3PT$_{suv}$(inverse) is much more time-consuming than the 3PT$_{suv}$ solver as shown in Table~\ref{tab:inverse}. 
In this experiment, we use GC-RANSAC~\cite{barath2017graph} without LO to show that 3PT$_{suv}$(inverse) solver is not practical. As such, we did not evaluate the 3PT$_{suv}$(inverse) solver in the main paper. 

\begin{table}[!ht]
	\begin{center}
        % \vspace{-2mm}
        % \resizebox{1.00\linewidth}{!}
        \resizebox{0.99\linewidth}{!}
        { 
	\begin{tabular}{ccccccc}
		\toprule
	 \multirow{2.5}{*}{{Depth}} &  \multirow{2.5}{*}{Method} & \multicolumn{5}{c}{\Phototourism }  \\
		\cmidrule(rl){3-7}
		 & &\ $\epsilon_{\M R}(^\circ)\downarrow$ & $\epsilon_{\M t}(^\circ)\downarrow$ & mAA($\M R$)$\uparrow$ & mAA($\M t$)$\uparrow$& $\tau (ms)\downarrow$ \   \\
        \midrule
        \multirow{2.0}{*}{DA V2~\cite{yang2024depthv2}}  & 3PT$_{suv}$ & {1.27} & {2.94} & {0.83} & {0.66} & 45.37  \\ 
         & 3PT$_{suv}$(inverse) & {1.28} & {3.02} & {0.83} & {0.65} & 194.77  \\ 
        \bottomrule
	\end{tabular}
        }
        \end{center}
        \vspace{-0.15in}
        \caption{Comparison between 3PT$_{suv}$ and 3PT$_{suv}$(inverse) using Depth anything V2~\cite{yang2024depthv2} on the \Phototourism dataset with GC-RANSAC~\cite{barath2017graph}.}
	\label{tab:inverse}
\end{table}

\subsection{Results for Fast 4PT Solvers}

Table~\ref{tab:eigen} shows that the relaxed eigenvalue solvers for the focal length problems are faster but give much worse results. Hence, we didn't use them in the real experiments.

\begin{table}[h]
	\begin{center}
        % \vspace{-2mm}
        % \resizebox{1.00\linewidth}{!}
        \resizebox{0.99\linewidth}{!}
        { 
	\begin{tabular}{ccccccccc}
		\toprule
	 \multirow{2.5}{*}{{Depth}} &  \multirow{2.5}{*}{Method} & \multicolumn{7}{c}{Phototourism} \\
		\cmidrule(rl){3-9} 
		 & &\ $\epsilon_{\M R}(^\circ)\downarrow$ & $\epsilon_{\M t}(^\circ)\downarrow$ & $\epsilon_{f}\downarrow$ & mAA($\M R$)$\uparrow$ & mAA($\M t$)$\uparrow$ & mAA($f$)$\uparrow$ & $\tau (ms)\downarrow$ \  \\
         \midrule
        \multirow{4.0}{*}{DA V2~\cite{yang2024depthv2}}  & 4PT$_{suv}f$ & {2.32} & {6.58} & {0.22} & {0.72} & {0.44} & {0.33} & 50.99  \\
         & 4PT$_{suv}f$(Eigen) & {5.17} & {17.25} & {0.30} & {0.52} & {0.22} & {0.21} & {8.18}  \\
         \cmidrule(rl){2-9} 
         & 4PT$_{suv}f_{1,2}$ & {5.78} & {17.37} & {0.26} & {0.48} & {0.20} & {0.23} & 54.27  \\
         & 4PT$_{suv}f_{1,2}$(Eigen) & {7.65} & {23.42} & {0.32} & {0.39} & {0.15} & {0.18} & {7.92}\\
		\bottomrule
	\end{tabular}
        }
        \end{center}
        \caption{Comparison between the focal length solvers shown in the main paper and the fast eigenvalue solutions inside GC-RANSAC~\cite{barath2017graph}.}
	\label{tab:eigen}
\end{table}

\subsection{More Results}

We provide more results for the three different cases including more datasets, RANSAC configurations, and additional solvers. Tables~\ref{tab:calib_eth_sm}-\ref{tab:calib_scannet_sm} show results for the calibrated case for the {{\fontfamily{cmtt}\selectfont ETH3D}}, \Phototourism and \ScanNet datasets respectively. We note that Mast3r~\cite{leroy2024grounding} with its non-linear optimization strategy is not included for the calibrated case, since the authors recommend using the 5PT~\cite{nister2004efficient} solver with RANSAC to obtain the poses instead.
Tables~\ref{tab:shared_eth_sm} and~\ref{tab:shared_scannet_sm} show results for shared focal length case for the \ETH and \ScanNet datasets. Tables~\ref{tab:var_eth_sm}-\ref{tab:var_scannet_sm} show retults for different uknown focal length case for the {{\fontfamily{cmtt}\selectfont ETH3D}}, \Phototourism and \ScanNet datasets respectively.

Tables~\ref{tab:calib_eth_sm}-\ref{tab:var_scannet_sm} include an alternative configuration of PoseLib~\cite{poselib} in which we use the reprojection error for scoring and LO denoted as R or its version with included shift denoted as R$_s$. Using this strategy generally does not produce improvements over using the Sampson error. However, we note that for \ScanNet using the reprojection error outperforms Sampson error when the estimated depth is accurate. This may be due to the fact that \ScanNet contains only indoor scenes, and thus depth estimates may be more reliable. %This further confirms our findings in the main paper that a hybrid strategy such as the one presented in \cite{yu2025relative} may be beneficial when considering depth.

We have also evaluated our proposed solvers 3PT$_{100}f$ and 3PT$_{100}f_{1,2}$ for the shared and different unknown focal cases respectively. 
These solvers assume zero shifts and known scales or same scales in both images and thus known scale ratio $s$.
On \Phototourism and \ETH they perform worse than alternatives. However, when evaluated on \ScanNet these solvers perform on par with solvers considering scale and shift. Additionally, solvers that do not model scale and shift can still produce reasonable results when using Mast3r's depth, as Mast3r inherently corrects depth scales based on multi-view information. This suggests that in some scenarios (such as indoor scenes) MDEs may provide depths for which scale and shift do not need to be considered.

For the case of different focal lengths on the \ScanNet dataset Mast3r~\cite{leroy2024grounding} with its optimization strategy achieves the best results. However, we show that these results can be surpassed when Mast3r matches are used in conjunction with MoGe~\cite{wang2024moge} for depth estimation. For this combination, the hybrid RANSAC strategy~\cite{yu2025relative} with either of the evaluated solvers yields better accuracy in both pose and focal length than Mast3r. We note that the runtime evaluation is fair, since for Mast3r runtime we do not include the inference time of the network which produces the matches.

\begin{table}[]
    \centering
    \resizebox{\linewidth}{!}{
}
    \caption{Results for the calibrated case on the \ETH dataset~\cite{schops2017multi}. Opt.: S, R, R$_s$ - PoseLib~\cite{poselib} implementation using Sampson error (S), reprojection error (R) or reprojection error with shift considered (R$_s$), H - hybrid RANSAC from~\cite{yu2025relative}, M - non-linear optimization used in~\cite{leroy2024grounding}.}
    \label{tab:calib_eth_sm}
\end{table}

\begin{table}[]
    \centering
        \resizebox{\linewidth}{!}{
%
}
    \caption{Results for the calibrated case on the \Phototourism dataset~\cite{Jin2020}. Opt.: S, R, R$_s$ - PoseLib~\cite{poselib} implementation using Sampson error (S), reprojection error (R) or reprojection error with shift considered (R$_s$), H - hybrid RANSAC from~\cite{yu2025relative}, M - non-linear optimization used in~\cite{leroy2024grounding}.}
    \label{tab:calib_pt_sm}
\end{table}

\begin{table}[]
    \centering
        \resizebox{\linewidth}{!}{
%
}
    \caption{Results for the calibrated case on the \
    \ScanNet dataset~\cite{dai2017scannet}. Opt.: S, R, R$_s$ - PoseLib~\cite{poselib} implementation using Sampson error (S), reprojection error (R) or reprojection error with shift considered (R$_s$), H - hybrid RANSAC from~\cite{yu2025relative}, M - non-linear optimization used in~\cite{leroy2024grounding}.}
    \label{tab:calib_scannet_sm}
\end{table}

\begin{table}
    \centering
        \resizebox{\linewidth}{!}{
%
}
    \caption{Results for the case of two cameras with shared unknown focal length on the \
    \ETH dataset~\cite{schops2017multi}. Opt.: S, R, R$_s$ - PoseLib~\cite{poselib} implementation using Sampson error (S), reprojection error (R) or reprojection error with shift considered (R$_s$), H - hybrid RANSAC from~\cite{yu2025relative}, M - non-linear optimization used in~\cite{leroy2024grounding}.}
    \label{tab:shared_eth_sm}
\end{table}

\begin{table}
    \centering
        \resizebox{\linewidth}{!}{
%
}
    \caption{Results for the case of two cameras with shared unknown focal length on the \
    \ScanNet dataset~\cite{dai2017scannet}. Opt.: S, R, R$_s$ - PoseLib~\cite{poselib} implementation using Sampson error (S), reprojection error (R) or reprojection error with shift considered (R$_s$), H - hybrid RANSAC from~\cite{yu2025relative}, M - non-linear optimization used in~\cite{leroy2024grounding}.}
    \label{tab:shared_scannet_sm}
\end{table}

\begin{table}
    \centering
        \resizebox{\linewidth}{!}{
%
}

    \caption{Results for the case of two cameras with different unknown focal lengths on the \
    \ETH dataset~\cite{schops2017multi}. Opt.: S, R, R$_s$ - PoseLib~\cite{poselib} implementation using Sampson error (S), reprojection error (R) or reprojection error with shift considered (R$_s$), H - hybrid RANSAC from~\cite{yu2025relative}, M - non-linear optimization used in~\cite{leroy2024grounding}.}
    \label{tab:var_eth_sm}
\end{table}

\begin{table}
    \centering
        \resizebox{\linewidth}{!}{
%
}
    \caption{Results for the case of two cameras with different unknown focal lengths on the \
    \Phototourism dataset~\cite{Jin2020}. Opt.: S, R, R$_s$ - PoseLib~\cite{poselib} implementation using Sampson error (S), reprojection error (R) or reprojection error with shift considered (R$_s$), H - hybrid RANSAC from~\cite{yu2025relative}, M - non-linear optimization used in~\cite{leroy2024grounding}.}
    \label{tab:var_pt_sm}
\end{table}

\begin{table}
    \centering
        \resizebox{\linewidth}{!}{
%
}
\end{tabular}}
    \caption{Results for the case of two cameras with different unknown focal lengths on the \
    \ScanNet dataset~\cite{dai2017scannet}. Opt.: S, R, R$_s$ - PoseLib~\cite{poselib} implementation using Sampson error (S), reprojection error (R) or reprojection error with shift considered (R$_s$), H - hybrid RANSAC from~\cite{yu2025relative}, M - non-linear optimization used in~\cite{leroy2024grounding}.}
    \label{tab:var_scannet_sm}
\end{table}

\end{document}